\documentclass[letterpaper, 10 pt, conference]{IEEEtran}  % Comment this line out if you need a4paper

\IEEEoverridecommandlockouts                              % This command is only needed if 
\usepackage{amsmath,amssymb,amsfonts}
\usepackage{algorithmic}
\usepackage{mathtools}
\usepackage{graphicx}
\usepackage{textcomp}
\usepackage{xcolor}
\usepackage{subcaption}
\usepackage{cuted}    % provides the strip environment (wide content)
\usepackage{capt-of}  % provides \captionof without touching IEEE captions
\usepackage{cleveref}
\usepackage{wrapfig}
\usepackage{adjustbox}   % to resize the table to fit within linewidth
\usepackage{booktabs}    % for \toprule, \midrule, \bottomrule
\usepackage[font=small]{caption}     % if you want to adjust caption style

\newcommand{\bb}{\mathbf} 
\begin{document}
\title{\LARGE \bf Cross-Modal Instructions for Robot Motion Generation
}

\author{William Barron$^{1,2}$ \and Xiaoxiang Dong$^{1,2}$ \and Matthew Johnson-Roberson$^{1,2}$ \and Weiming Zhi$^{1,2,3,*}$
\thanks{$^{*}$email: {\tt\small Weiming.Zhi@sydney.edu.au}.}%
\thanks{$^{1}$ College of Connected Computing, Vanderbilt University, TN, USA}
\thanks{$^{2}$ Robotics Institute, Carnegie Mellon University, PA, USA}
\thanks{$^{3}$ School of Computer Science, The University of Sydney, Australia.}}
\maketitle
\begin{strip}
\centering
\includegraphics[width=0.95\textwidth]{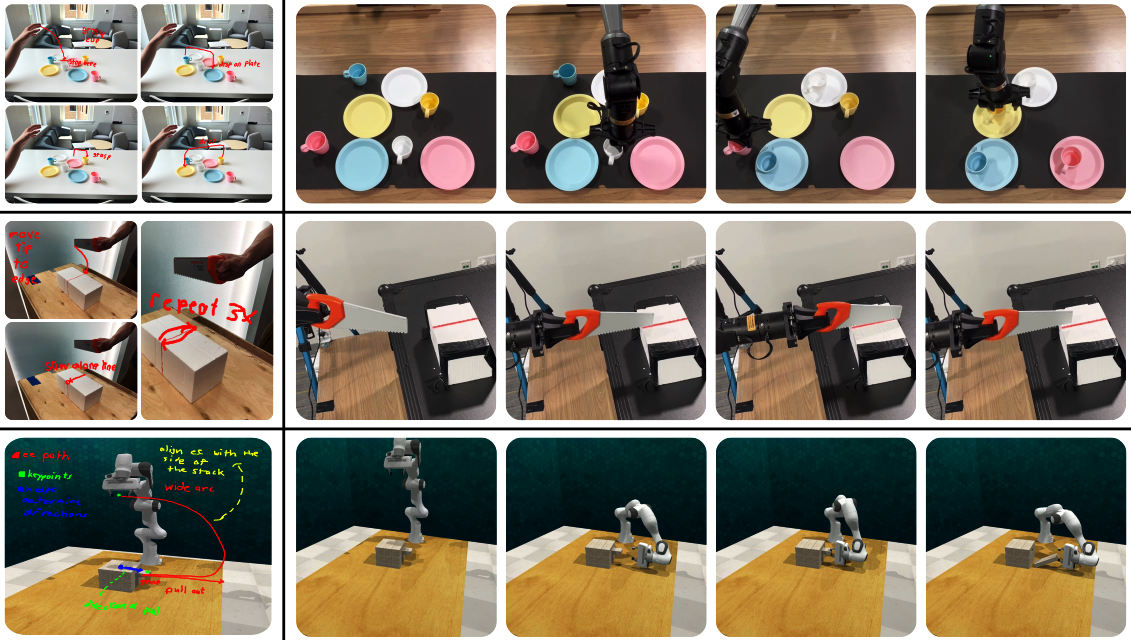}
\captionof{figure}{We enable robots to interpret \emph{cross-modal instructions}, in the form of rough sketches and textual labels (shown on the left). Subsequent motion can be generated and generalized to novel setups and environments (shown on the right).}
\label{fig:teaser}
\vspace{-1.25em}
\end{strip}

\begin{abstract}
Teaching robots novel behaviors typically requires motion demonstrations via teleoperation or kinaesthetic teaching, that is, physically guiding the robot. While recent work has explored using human sketches to specify desired behaviors, data collection remains cumbersome, and demonstration datasets are difficult to scale. In this paper, we introduce an alternative paradigm, Learning from Cross-Modal Instructions, where robots are shaped by demonstrations in the form of rough annotations, which can contain free-form text labels, and are used in lieu of physical motion. We introduce the \emph{CrossInstruct} framework, which integrates cross-modal instructions as examples into the context input to a foundational vision–language model (VLM). The VLM then iteratively queries a smaller, fine-tuned model, and synthesizes the desired motion over multiple 2D views. These are then subsequently fused into a coherent distribution over 3D motion trajectories in the robot's workspace. By incorporating the reasoning of the large VLM with a fine-grained pointing model, CrossInstruct produces executable robot behaviors that generalize beyond the environment of in the limited set of instruction examples. We then introduce a downstream reinforcement learning pipeline that leverages CrossInstruct outputs to efficiently learn policies to complete fine-grained tasks. We rigorously evaluate CrossInstruct on benchmark simulation tasks and real hardware, demonstrating effectiveness without additional fine-tuning and providing a strong initialization for policies subsequently refined via reinforcement learning.
\end{abstract}

\section{Introduction}
Imitation learning, also known as Learning from Demonstration, is a dominant paradigm for teaching robots new skills \cite{ravichandar2020recent}. Imitation learning requires the collection of a dataset of demonstrations, which are generally provided by a human user teleoperating the robot via a remote controller or by kinesthetic teaching, i.e. physically handling the robot. In both of these cases, collecting a sufficient number of demonstrations for even marginal levels of generalization is challenging. There have also been efforts to reduce the human burden of providing demonstrations, including hardware adaptations \cite{chi2024universal}, and using human sketches as a demonstration interface \cite{diagrammaticlearning,yu2025sketch,mehta2025l2d2,singh2025varp, Spatial_diag}. Humans have a remarkable ability to adapt the same skill to different environments and setups, even after observing one or two demonstrations. However, imitation learning is far less demonstration-efficient, and it can be challenging to collect a sufficient number of demonstrations to generalize behavior when the scene changes. In this work, we aim to circumvent the cumbersome, large-scale human input required for imitation learning.

We introduce Learning from Cross-modal Instructions as an alternative paradigm to shape robot behavior. In the proposed method, human operators provide \emph{cross-modal instructions}, which are free-form sketches and textual labels on an image of an operating environment. The robot infers the intended behavior from these inputs and generalizes execution across varying setups. To learn from cross-modal instructions, we propose the CrossInstruct framework. \Cref{fig:teaser} illustrates several examples in both simulation and the real world with cross-modal instructions and subsequent robot behavior generated by CrossInstruct. CrossInstruct takes cross-modal instructions as \emph{in-context learning examples} \cite{dong-etal-2024-survey} for a large VLM. The large model performs high-level task reasoning and delegates pixel-level keypoint localization to a smaller VLM fine-tuned for 2D pointing, which returns precise coordinates for task-critical features identified by the large model. The hierarchical coupling between the large reasoning VLM with the smaller, fine-tuned model enables the efficient and accurate identification of task-relevant keypoints. These are then used to produce motion trajectory sketches over a small set of multi-view images of the setup. The resulting 2D trajectories are lifted into the robot’s workspace to yield a distribution over 3D trajectories. Combined with end-effector orientations and gripper actions generated by the reasoning model, this produces executable motion sequences.

%We seek the robot to interpret the desired behavior from the cross-modal instructions and generalize the robot's behavior to complete the task under varying setups and environments.

%After the trajectory sketches over multiple views are produced, CrossInstruct then casts these 2D trajectories into the robot's workspace to arrive at a distribution of 3D trajectories. Combined with reasoned gripper orientations and actions, CrossInstruct then produces executable robot motion sequences. 

We demonstrate the capabilities of CrossInstruct to shape behavior to complete and generalize across a wide range of tasks, in both simulation and in the real world, without additional fine-tuning. Beyond directly executing the produced motions, we can enable the robots to achieve an even higher level of precision by joining CrossInstruct with a downstream reinforcement learning pipeline. Concretely, the technical contributions in this paper include: 
\begin{itemize}
\item The Learning from Cross-Modal Instructions paradigm, circumventing the need for physical demonstrations;
\item The CrossInstruct framework to leverage cross-modal instructions in an in-context fashion to synthesize robot behavior. This is achieved by a hierarchical coupling of two VLMs and using raycasting to fuse multi-view 2D sketches into executable robot motion;
\item Evaluations on benchmark tasks, demonstrating robustness of CrossInstruct out of the box in both simulation and the real world, as well as leveraging the distribution of trajectories to enable efficient reinforcement learning.
\end{itemize}

\section{Related Work}
\textbf{Robot Motion Generation:}
Generating movements for robots is a critical area of study in robotics. Classical approaches formulate a motion planning problem \cite{Motion_planning, CHOMP, PDMP}, where desired goals and constraints are specified \emph{a priori}. Another popular paradigm, imitation learning or learning from demonstrations \cite{ravichandar2020recent}, circumvents the specification of ``how'' the motion should be, by mimicking a set of shown robot trajectory demonstrations. Various approaches have been proposed to formulate mimicking as a deterministic or probabilistic supervised learning problem to learn actions \cite{promp, implicitBC}, predict high-level way-points \cite{GeoFab_gloabL_opt}, or as a \emph{inverse} reinforcement learning problem \cite{inv_rl}.
 
\textbf{Sketches for Robot Learning:}
Providing robot trajectories as demonstrations often requires careful tele-operation or cumbersome kinesthetic teaching. Sketches over images have emerged as a lightweight channel for specifying motion. These include techniques for directly interpreting sketches of precise desired robot motion \cite{diagrammaticlearning, frc-sketch-moma, periodic}. Others translate sketches into 3D demonstrations that bootstrap data collection and reinforcement learning \cite{yu2025sketchtoskill}. A related direction conditions imitation policies directly on sketches, where drawings of desired outcomes are provided as goals, guiding behavior \cite{pmlr-v270-sundaresan25a}. Sketching has emerged as an impactful interface for specifying fine-grained details for robot motions. 

\textbf{Large Pre-trained Models in Robotics:}
Large language and vision–language models have been used as task planners, decomposing natural-language goals into perception queries and motion objectives \cite{yang2025vlmtamp}. Earlier work revolves around grounding language in action, these include Perceiver-Actor \cite{shridhar2022peract} and SayCan \cite{Ahn2022DoAI}. Many approaches construct 3D value maps optimized by motion planners \cite{pmlr-v229-huang23b}, while others output coarse sketches refined by low-level controllers \cite{li2025hamster}. In these cases, models remain high-level reasoners, with trajectory generation handled by separate planning or control modules. Vision–language models have also been employed to structure perception through semantic cues. Examples include open-vocabulary detectors that produce voxelized affordance fields optimized by planners \cite{pmlr-v229-huang23b}, and hierarchical structures for reasoning \cite{fangandliu2024moka}. Instruction-tuned VLMs have also been applied for pixel-level pointing, producing 2D keypoints or affordances without explicit control code or volumetric fields \cite{pmlr-v270-yuan25c}. These strategies highlight the utility of VLMs in providing precise and semantically rich cues for downstream planning.

\begin{figure}[t]
    \centering
    \framebox{\includegraphics[width=0.49\linewidth]{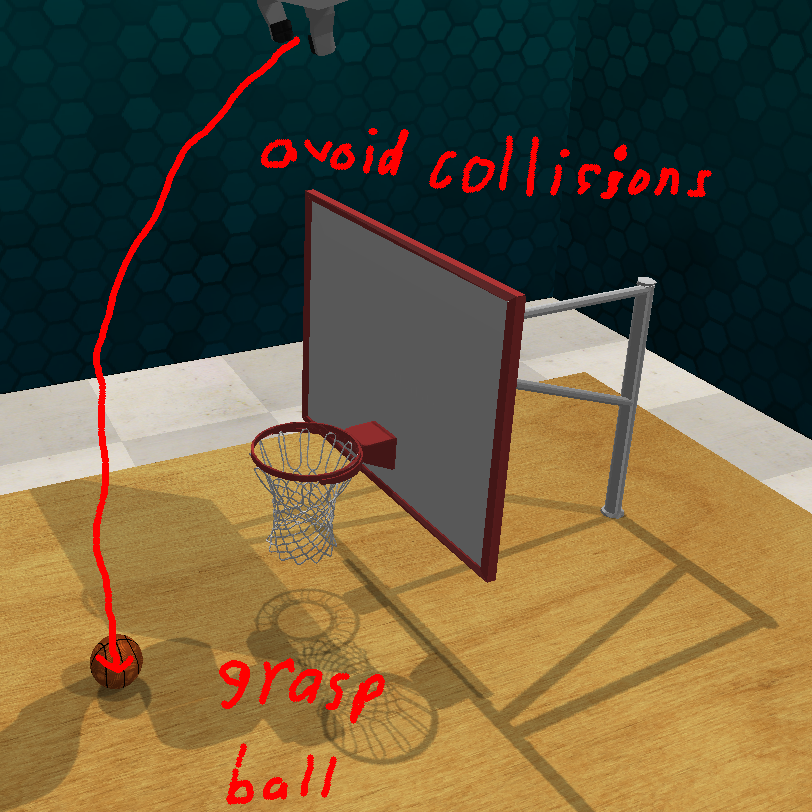}}%
    \framebox{\includegraphics[width=0.49\linewidth]{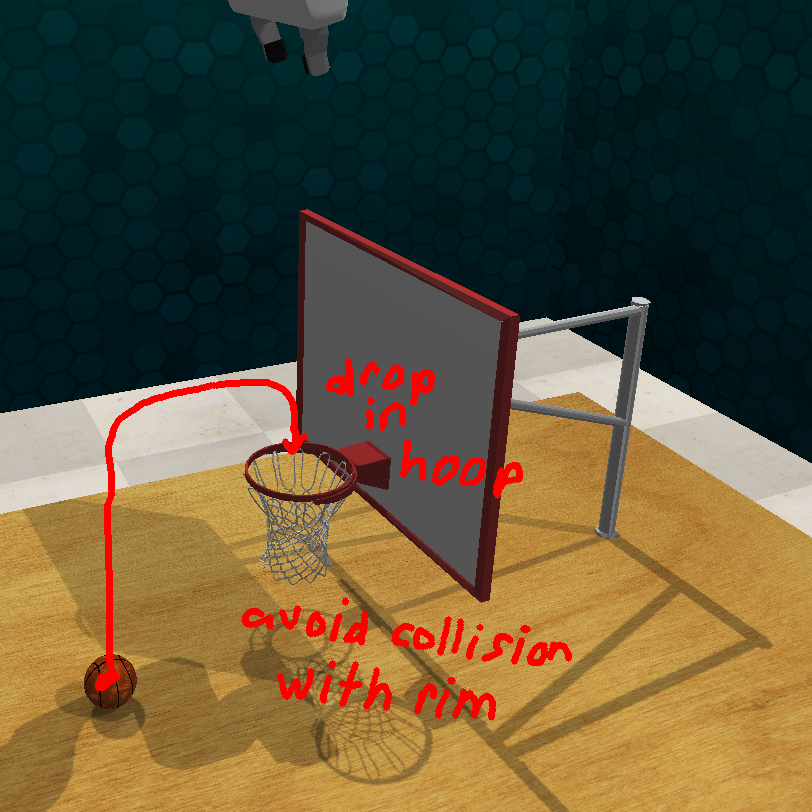}}
    \caption{Example cross-modal instructions which a human can specify to complete the \emph{Basketball in Hoop} problem in the RLBench benchmark \cite{RLBench}.}
    \label{fig:example_instructions}
\end{figure}

\section{Learning From Cross-Modal Instructions}

We consider a robot operating in a workspace $\mathcal{W} \subseteq \mathbb{R}^3$, 
where the goal is to generate executable motion trajectories conditioned on 
\emph{cross-modal instructions}. 
\begin{figure*}[t]
    \centering
    \begin{subfigure}[b]{0.595\textwidth}
    \includegraphics[width=\linewidth]{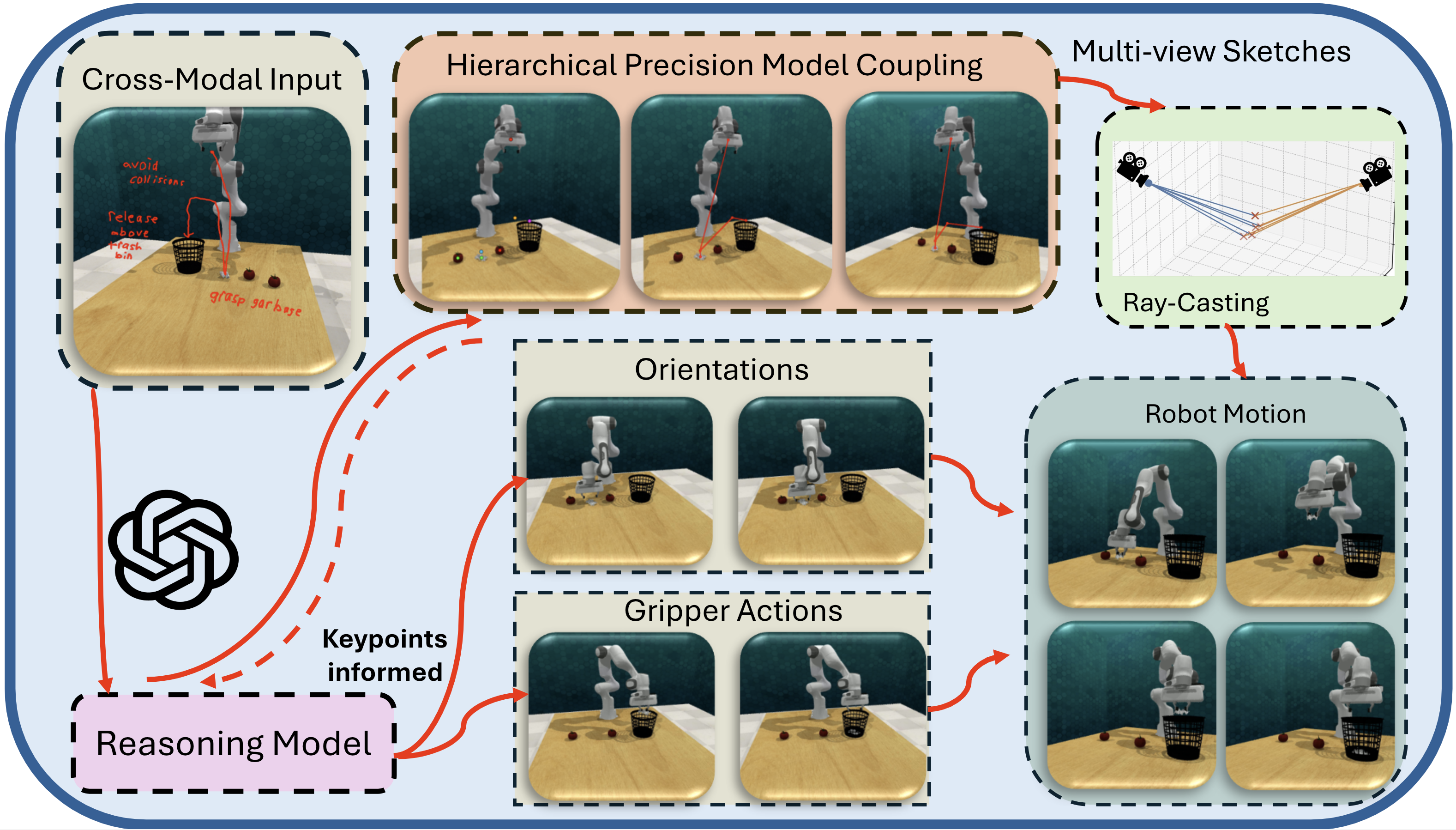}
    \end{subfigure}
    \begin{subfigure}[b]{0.395\textwidth}
    \includegraphics[width=\linewidth]{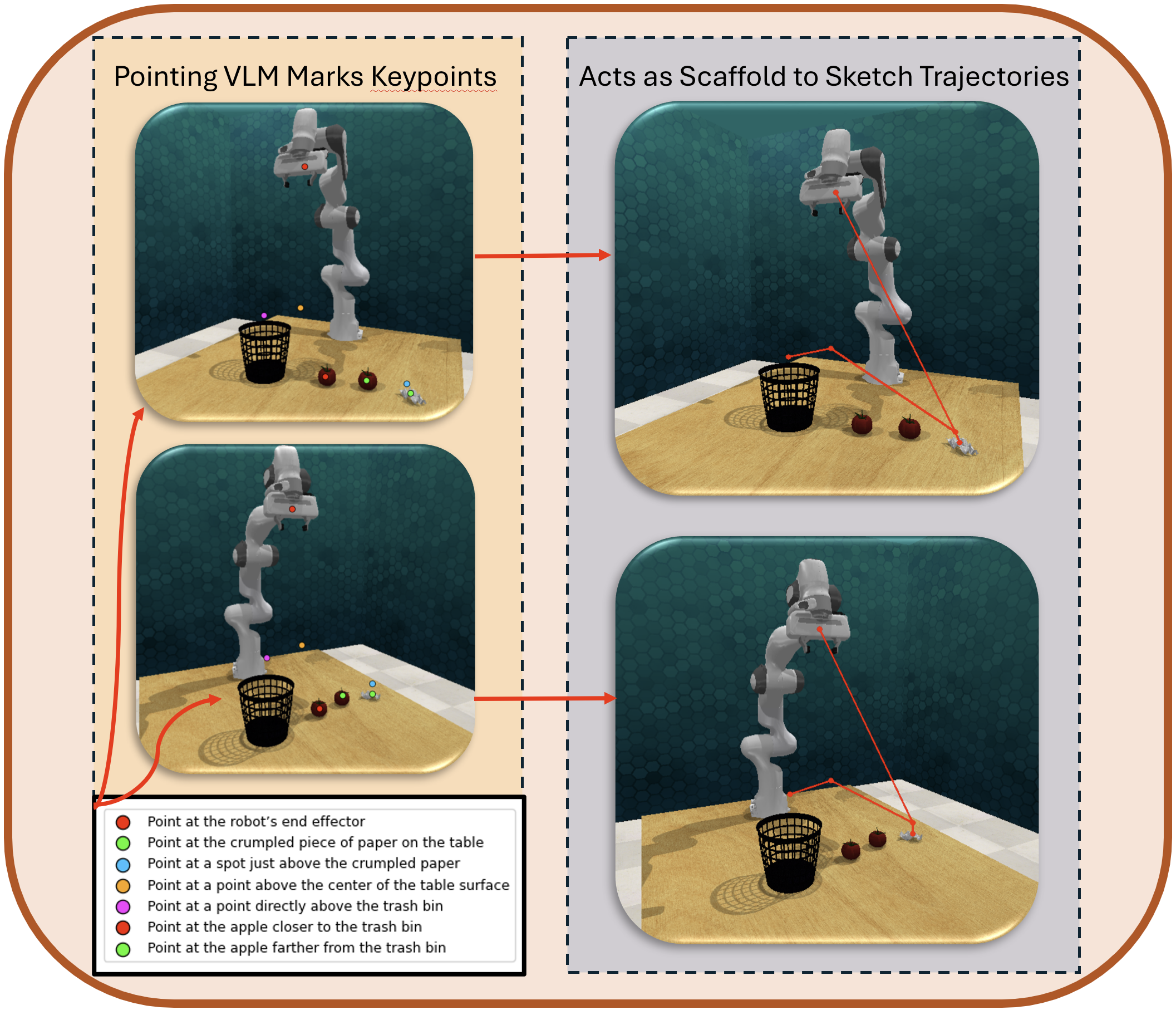}
    \end{subfigure}
    \caption{An overview of CrossInstruct (left) with an example of precise pointing of keypoints (right). The hierarchical precision model coupling module enables the reasoning model to leverage a smaller fine-tuned VLM to precisely identify relevant keypoints, which then guide robot end-effector motion.}
    \label{fig:Overviews}
\end{figure*}
A cross-modal instruction is defined as a tuple
\begin{align}
\mathcal{I} = \{ I, S, T \},
\end{align}
where $I$ is an image of a relevant scene, 
$S$ is a set of free-form sketches provided by a human operator 
(e.g., lines, arrows, or annotated paths over $I$), 
and $T$ is an optional set of textual information describing constraints and the intended action, which can be directly handwritten over the image. An example of such cross-modal instructions, including arrow notations and text, are given in \cref{fig:example_instructions}.

The goal is for the robot to generalize the behavior captured in the instructions to a new environment or setup. Given a tuple of multi-view images of the new scene, $\mathcal{V}$, along with their poses, $\mathcal{P}$, the robot must map $\mathcal{I}$ to a \emph{distribution over motion trajectories} 
in the workspace:
\begin{align}
p(\tau \mid \mathcal{I},\mathcal{V},\mathcal{P}), 
\quad \tau = \{ (x_t, R_t, g_t) \}_{t=1}^H ,
\end{align}
where $x_t \in \mathbb{R}^3$ is the position of the end-effector at time $t$, 
$R_t \in SO(3)$ is the orientation, $g_t \in \{0,1\}$ encodes the open or closed grip state and $H$ is the trajectory horizon. During execution, the mean of the trajectory distribution $\mathcal{E}(\tau)$ is tracked, while the entire distribution can be used to generate varying trajectories in order to adequately train downstream reinforcement learning policies.

The learning problem is thus, given cross-modal instructions $\mathcal{I}$, to
infer a trajectory distribution $p(\tau \mid \mathcal{I},\mathcal{V},\mathcal{P})$ that generalizes across environments. Unlike imitation learning, where supervision comes from expert trajectories
$\mathcal{D}_{\text{IL}} = \{ \tau^{(i)} \}_{i=1}^N$,
our supervision consists only of weak, symbolic, or high-level annotations
$\{ \mathcal{I}^{(i)} \}_{i=1}^M$.
The challenge is to \emph{bridge modalities}, from free-form sketches to executable motion trajectories.

%\section{Vision-Language Sketch Motion}
%\subsection{Problem Setup and Overview}

\begin{figure}[t]
    \centering
    \includegraphics[width=\linewidth]{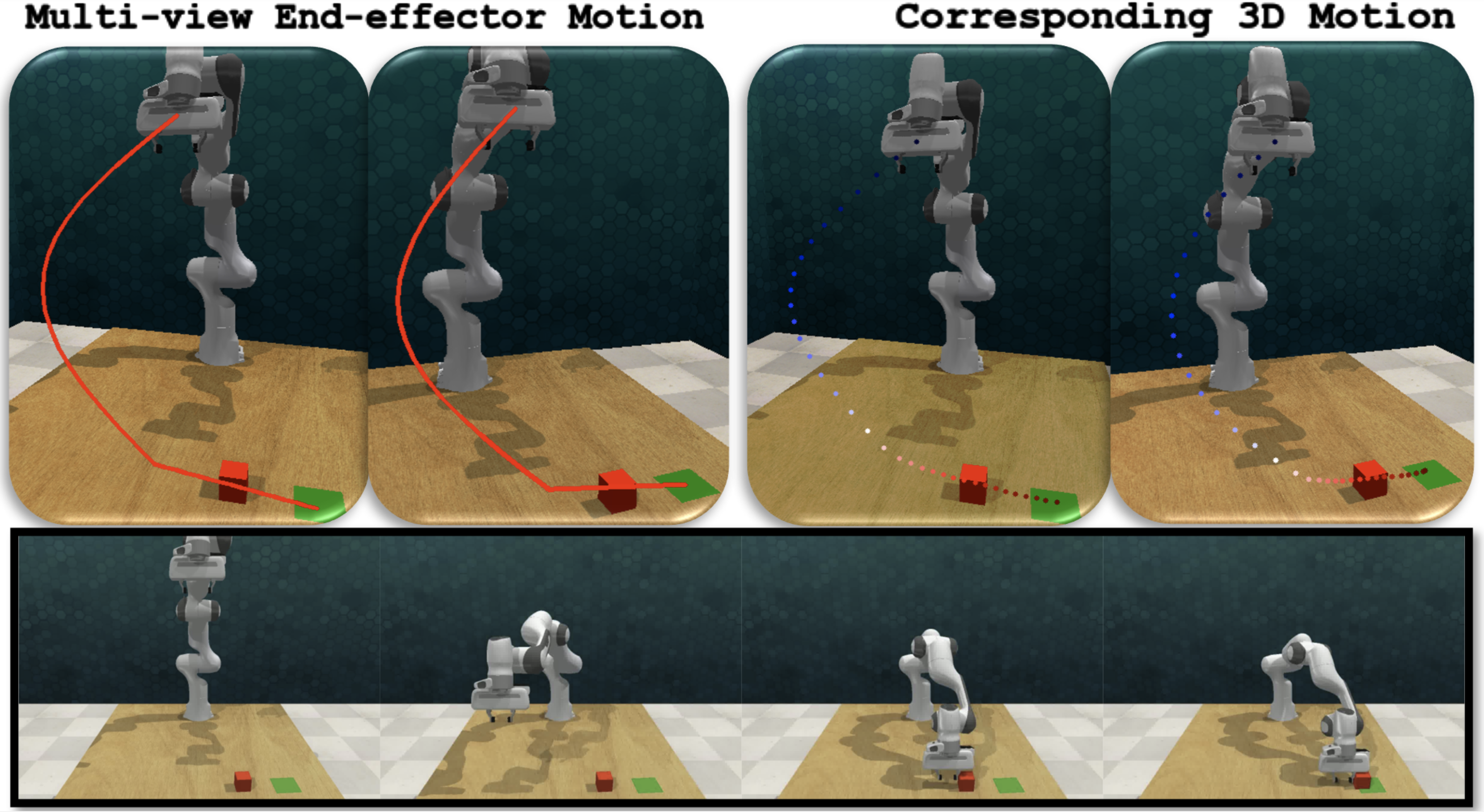}
    \caption{CrossInstruct generates 2D trajectories over multi-view images (in red), which are subsequently fused into a coherent 3D trajectory (waypoints in blue-red color gradient shown). This can then be rolled out to slide the block to the target.}
    \label{fig:2d_to_3d}
\end{figure}

\section{CrossInstruct Framework}\label{sec: CrossInstruct}

The goal of CrossInstruct is to transform human-provided \emph{cross-modal instructions}, sketches $S$ and textual labels $T$ defined over scene images $I$, into executable robot motion trajectories in $\mathbb{R}^3$. This problem requires both high-level semantic interpretation of human intent and precise spatial grounding in the robot’s workspace. Our framework addresses this through a \textbf{hierarchical precision coupling} between a large reasoning vision–language model (VLM) and a lightweight, fine-tuned pointing model. The reasoning VLM primarily operates at the symbolic level (task identification, task decomposition, intent parsing, trajectory refinement), while the fine-tuned model operates at the perceptual level (pixel-localization of keypoints). An overview of our CrossInstruct framework is given in \Cref{fig:Overviews}.

\subsection{Hierarchical Precision Coupling Module}

The transformation from symbolic, cross-modal instructions $\mathcal{I} = \{ I, S, T \}$ to grounded motion trajectories requires reasoning across two levels of abstraction: high-level semantic interpretation of task intent and low-level pixel precision in image space. To achieve this, we introduce a hierarchical precision coupling module that tightly integrates a large reasoning VLM $\mathcal{R}$ with a smaller, fine-tuned pointing model $\mathcal{G}$. The hierarchical structure allows $\mathcal{R}$ to maintain a global task context, while delegating precise spatial localization, or \emph{pointing}, to $\mathcal{G}$, which is specifically trained to resolve ambiguous references in concrete pixel coordinates. In this work, our large reasoning VLM is OpenAI o3 \cite{o3systemcard2025}, and the fine-tuned pointing VLM is Molmo \cite{molmo2024}. 

Given an instruction $\mathcal{I}$ and a tuple of multi-view images $\mathcal{V} = \{ I_1, I_2 \}$, the reasoning model $\mathcal{R}$ first generates a set of $N$ semantic keypoint descriptors,
\begin{align}
\mathcal{K} = \{ k_i \}_{i=1}^N, \qquad k_i = (\ell_i, \alpha_i),
\end{align}
where each $\ell_i$ is a natural language label and $\alpha_i$ denotes auxiliary text-based metadata and additional details, such as proximity descriptions or qualitative constraints. These descriptors are context-dependent and reflect $\mathcal{R}$’s global reasoning over both the free-form sketches contained in the instructions, and scene structure contained in the two multi-view images. However, the keypoints are not yet linked to the pixel locations.

The keypoint descriptors are then passed to a fine-tuned pointing model $\mathcal{G}$, which receives each descriptor $\ell_i$ together with the images $I_1$ and $I_2$, and predicts a concrete pixel coordinate per keypoint and per multi-view image,
\begin{align}
\{u_{i,m}^{(t)}, v_{i,m}^{(t)}\} = \mathcal{G}(\ell_i \mid Im), && m=\{1,2\}, i=1,\ldots, N.
\end{align}
yielding a set of 2D keypoints across views. As $\mathcal{G}$ is trained for pixel-wise pointing within images, we obtain a set of precise keypoints indicating the most salient features relevant to task completion. 

The set of keypoints is then added to the context maintained by the larger reasoning model $\mathcal{R}$. The context now contains the original keypoint descriptors $\mathcal{K}$, with precise locations $\{\{u_{i,m},v_{i,m}\}_{i=1}^{N}\}_{m=1}^{2}$. Based on this updated context, $\mathcal{R}$ produces two 2D trajectories over each multi-view image. That is,
\begin{align}
\xi_{1}, \xi_{2} = \mathcal{R}(\mathcal{I}, \mathcal{V}, \mathcal{K},\{(u_{i,m}^{(t)}, v_{i,m}^{(t)})\}),
\end{align}
where $\xi_{1}$ and $\xi_{2}$ are equal-length pixels over the two views.

\begin{figure}[t]
    \centering
    \includegraphics[width=0.1975\linewidth]{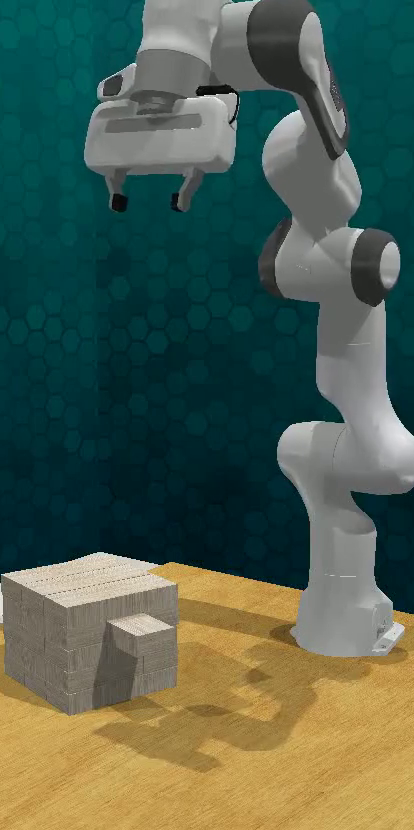}%
    \includegraphics[width=0.1975\linewidth]{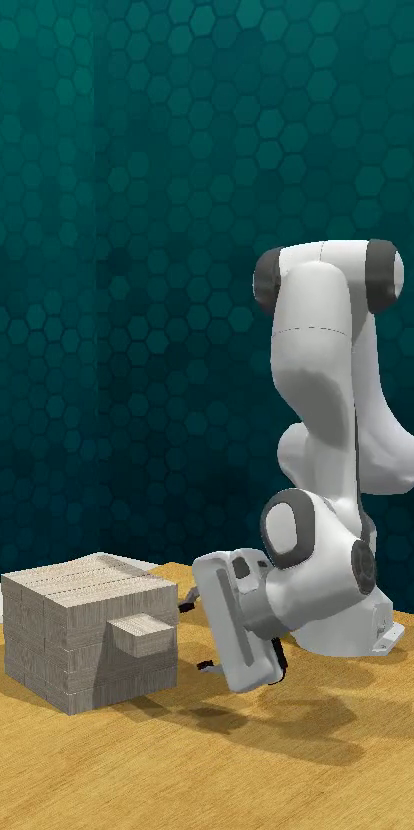}%
    \includegraphics[width=0.1975\linewidth]{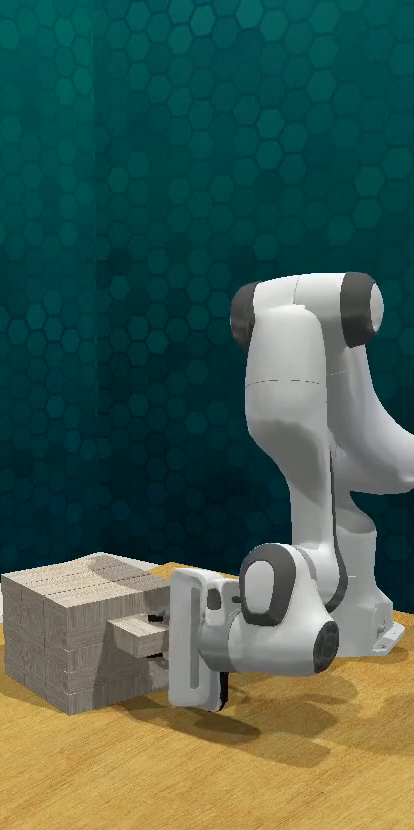}%
    \includegraphics[width=0.1975\linewidth]{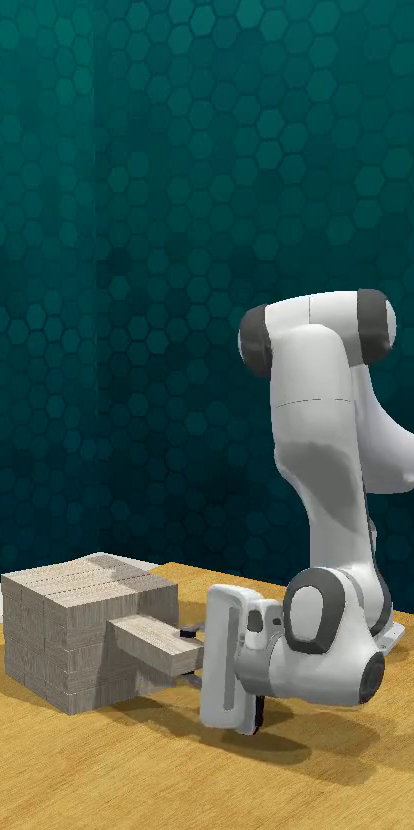}%
    \includegraphics[width=0.1975\linewidth]{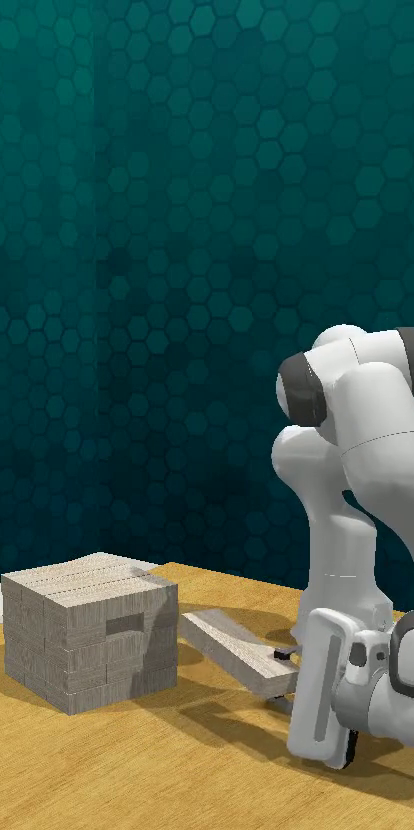}%
    \caption{CrossInstruct uses the accurately pointed keypoints to enable accurate reasoning of orientation. Here we see an example of the robot pulling a Jenga block, which requires an accurate movement direction.}
    \label{fig:orientation}
\end{figure}

This design highlights two key principles. First, the hierarchy enables each model to operate at its natural level of abstraction: $\mathcal{R}$ interprets high-level symbolic cues and manages task context, whereas $\mathcal{G}$ delivers precise geometric observations. Example end-effector trajectories are shown in \cref{fig:2d_to_3d}, in the left subfigure. These are then ray-casted into the 3D workspace to estimate a 3D trajectory of end-effector motion. The right subfigure in \cref{fig:2d_to_3d} shows the motion trajectory in 3D visualised as a sequence of waypoints. The reasoning model then takes the 3D trajectory as an in-context example and specifies end-effector orientation and gripper actions to arrive at a desired trajectory to follow, $\tau$. \Cref{fig:orientation} shows the robot behavior generated when the 3D end-effector trajectory, orientation, and gripper actions are combined. In the next section, we shall elaborate on how to obtain estimates of 3D trajectories from $\xi_{1}$, $\xi_{2}$.

\subsection{Multi-View Lifting via Ray Casting}
\begin{wrapfigure}{L}{0.38\linewidth}
\centering
\fbox{\includegraphics[width=\linewidth]{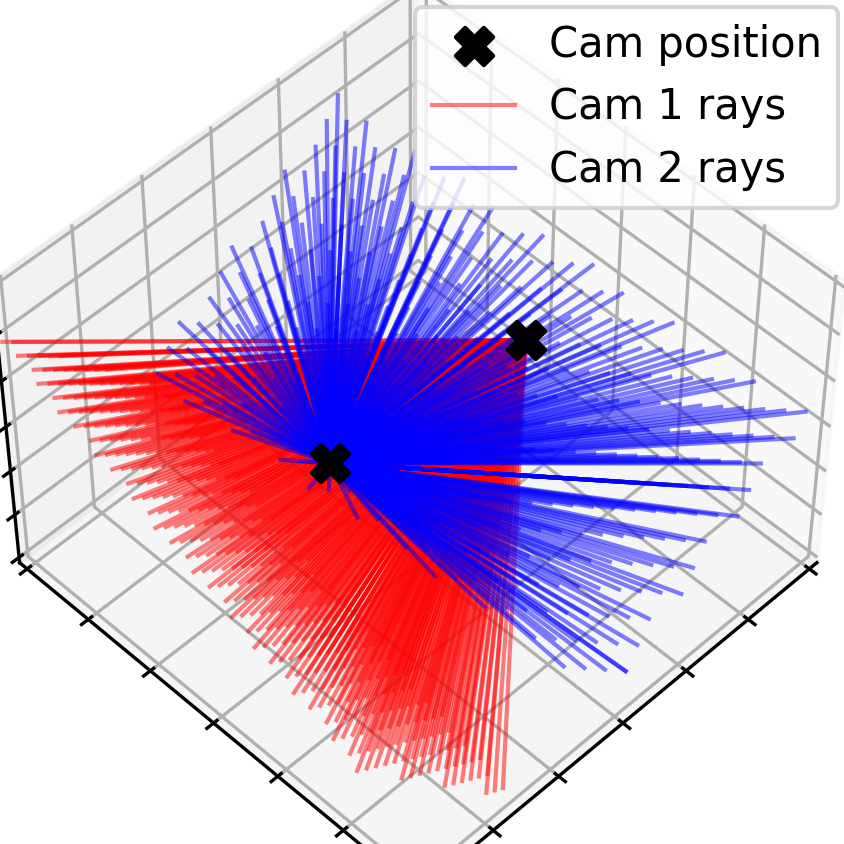}}
\caption{Intersecting rays from different poses. }\label{fig:example_ray}
\end{wrapfigure}
The trajectories $\xi_1, \xi_2$ produced by the reasoning model define continuous curves in pixel space. To embed them into the 3D workspace $\mathcal{W} \subseteq \mathbb{R}^3$, we lift them through calibrated multi-view geometry. The camera poses can either be determined via external tags or prior calibration \cite{jcr, hand-eye, Sim_pose}. We interpret each 2D trajectory $\xi_m$ as defining a time-indexed probability distribution over pixels, $p(u_m, v_m \mid t)$, for view $m \in \{1,2\}$. Specifically, for a given time $t$, the pixel coordinate along the curve, $(u_{t,m}, v_{t,m}) = \xi_m(t)$, defines the mean of a Gaussian density in image space:
\begin{align}
p(u_m, v_m \mid t) = \mathcal{N}\big((u_m, v_m) \,\big|\, \xi_m(t), \Sigma_m \big),
\end{align}
where $\Sigma_m$ is an assumed covariance that captures uncertainty, which can be predefined. Uncertainty in the robot pipeline provides additional robustness \cite{senanayake2024role}. Thus, rather than a single pixel, each waypoint induces a distributional neighborhood in the view space. We adopt a standard pinhole camera model to project these pixel densities into the 3D workspace, by casting rays through each camera view and sampling to find where the rays intersect \cite{ray_tracing_course}. As illustrated in \Cref{fig:example_ray}, intersecting viewing rays from different calibrated poses localize a 3D waypoint that is consistent across both per-view selections. For view $m$, a pixel $(u_m,v_m)$ corresponds to a ray
\begin{align}
f_{r}(u_m,v_m, d) = \bb{o}^{m} + \bb{\omega}(u_m,v_m)\, d, 
&& d \in [d_{\text{near}}, d_{\text{far}}],
\end{align}
where $\bb{o}^{m}$ is the camera origin, $\bb{\omega}(u_m,v_m)$ is the ray direction derived from the intrinsic parameters of the camera, and $d$ is the depth parameter. For a fixed time $t$, the 3D region consistent with the 2D Gaussian density is
\begin{align}
R^m_t = \{ \bb{x} \in \mathbb{R}^3| 
\bb{x} = f_r(u_m,v_m, d),\;
p(u_m,v_m \mid t) \geq \epsilon\}.
\end{align}
The 3D trajectory support at time $t$ is then approximated by the intersection of these ray-cast regions from both views, $\mathcal{R}_t = R^1_t \cap R^2_t$. In practice, we sample points along $d$ and $(u_m,v_m)$ in both views, and retain pairs whose cross-view projections are within a tolerance $\delta$. This produces a set of spatial samples
$\mathcal{S}_t = \{ \bb{x}_i \}_{i=1}^{n_t}$, which approximate the feasible 3D positions of the end-effector at time $t$. Finally, we fit a concise Gaussian distribution to these samples to model the time-conditioned waypoint distribution:
\begin{align}
p(x_t \mid t) = \mathcal{N}(x_t \mid \mu_t, \Sigma_t),
\end{align}
where $\mu_t$ is the mean of $\mathcal{S}_t$ and $\Sigma_t$ its empirical covariance. Repeating this across all $t=1,\ldots,H$ yields a trajectory distribution
\begin{align}
p(\tau \mid \xi_1, \xi_2, \mathcal{P}) = \prod_{t=1}^H p(x_t \mid t).
\end{align}
The direction of the ray $\bb{\omega}(u_m,v_m)$ is obtained by normalizing the coordinates of the pixels with the intrinsics of the camera of each view, $K_m$, and rotating into the world frame with $R_m^\top$. The mean trajectory $\mathbb{E}[\tau] = \{ \mu_t \}_{t=1}^H$, combined with the end-effector orientation, can be executed directly. Otherwise, trajectories can be generated for further fine-tuning, where the covariance of the trajectory distribution is considered when generating samples.  

\subsection{Policy Execution and Reinforcement Learning Refinement}\label{subsec:refine}

The synthesized trajectory $\mathbb{E}[\tau]$ can be executed directly as an open-loop plan $\pi_\tau$. However, in general, $\tau$ provides a source of demonstration data that is used both to initialize the policy and to regularize learning through loss of behavior cloning (BC). This allows the system to bootstrap reinforcement learning with informative samples that remain aligned with the cross-modal instructions.

To this end, we adopt Twin Delayed DDPG (TD3) \cite{Fujimoto2018AddressingFA} augmented with Behavior Cloning. We first construct a data set of trajectories, which can be achieved by executing $\tau$ in simulation, yielding a set of state–action pairs. The robot manipulator's state is described by the end-effector pose and a binary gripper state, and the action is given by the difference in end-effector pose and gripper state. By sampling $p(\tau|\xi_{1}, \xi_{2},\mathcal{P})$ over a set of problem setups, we arrive at the dataset,
\begin{align}
\mathcal{D} = \{ (s_t, a_t) \}_{t=1}^H.
\end{align}
This dataset serves two functions. First, $\mathcal{D}$ is used to pretrain the policy $\pi$ by cloning behavior, providing a strong initialization that accelerates learning. Second, $\mathcal{D}$ is retained throughout reinforcement learning, where it contributes to the BC regularization term in the actor loss. The TD3+BC actor loss is
\begin{align}
\mathcal{L}_{\text{actor}} =& \lambda \, \mathbb{E}_{(s,a)\sim \mathcal{D}}[\| \pi(s) \nonumber  - a \|^2] \nonumber\\
&- (1-\lambda)\,\mathbb{E}_s[Q(s,\pi(s))],
\end{align}
where $\pi$ is the policy, $Q$ is the critic, and $\lambda \in [0,1]$ controls the balance between imitation and value-based optimization. 

By construction, this loss ensures that the learned policy remains close to the synthesized trajectory while still improving beyond it through critic-guided updates. During training, the agent converges to a closed-loop policy $\pi^*$ that is robust to disturbances, perception noise, and distributional shifts, while remaining grounded in the weak supervision provided by the original instruction. The resulting policy is capable of tackling problems where fine-grained motion is required.

\begin{figure}[t]
    \centering
    \includegraphics[width=1.0\linewidth]{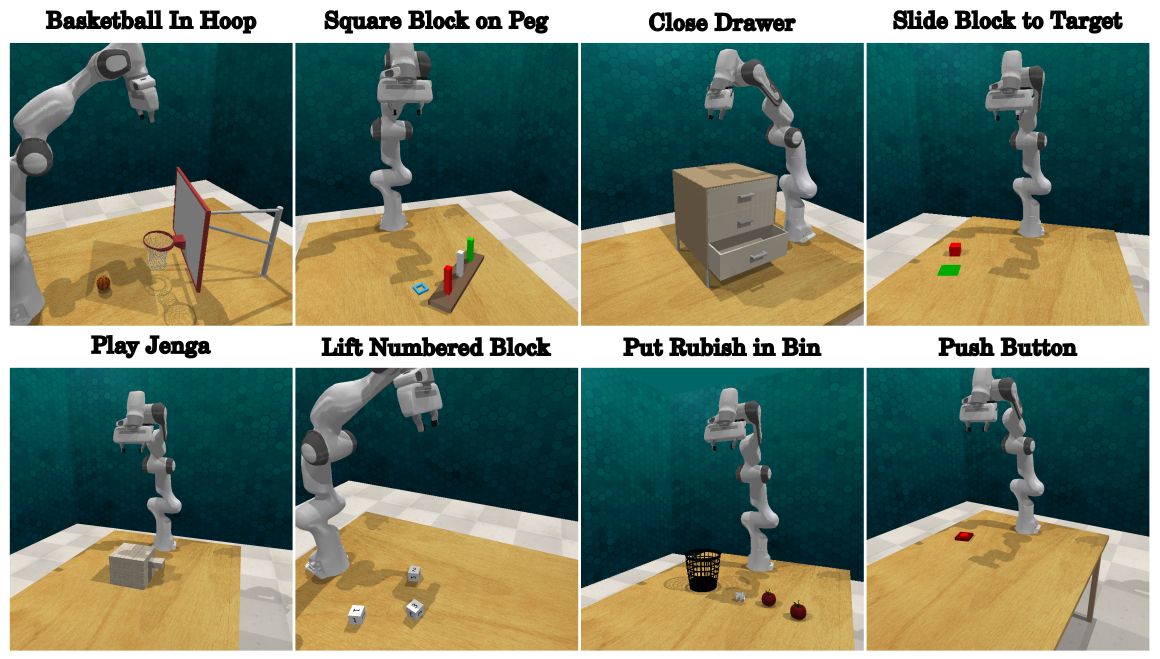}
    \caption{Representative tasks from the RLBench \cite{RLBench} benchmark used in our evaluations.}
    \label{fig:Rep_images}
\end{figure}

\section{Empirical Evaluations}
We rigorously evaluate CrossInstruct in both simulated and real-world robot learning tasks. Our experiments are designed to address the following research questions: 
\begin{itemize}
    \item How does CrossInstruct, which integrates a hierarchical precision-coupling module, compare against conventional VLM reasoning baselines and pure reinforcement learning (RL) baselines?
    \item Can CrossInstruct generalize to real world setups where the demonstrations are collected in environments that differ substantially, in both appearance and embodiment, from the robot’s execution environment?
    \item Can CrossInstruct be used to generate robot motion trajectories to warm-start reinforcement learning (RL), and what are the benefits of the downstream fine-tuning in terms of convergence for the RL policy? 
\end{itemize}

\subsection{Experimental Setup}
\paragraph{Simulation} We conduct controlled experiments in simulation using the Robot Learning Benchmark (RLBench) \cite{RLBench}. RLBench provides a diverse suite of goal-directed manipulation tasks, each equipped with programmatic success checks and automatic scene randomization. These properties allow us to perform rigorous seed-based evaluations of both perception and control under varying conditions. The simulator is configured with a 7-degree-of-freedom (DoF) Franka Emika Panda arm mounted on a tabletop workspace. All policies are evaluated over 20 held-out random seeds per task, which are strictly disjoint from those used for demonstration collection.  

\paragraph{Hardware} For real-world experiments, we deploy CrossInstruct on a 6-DoF AgileX manipulator with a parallel-jaw gripper. RGB images are captured from a fixed overhead camera. Importantly, the manipulator differs in morphology, kinematics, and appearance from the Franka Panda used in simulation.  

\paragraph{Tasks} The selected RLBench tasks span both contact-rich and precision-sensitive skills: \textit{Basketball-in-Hoop, Close Drawer, Insert Square Block on Peg, Lift Numbered Block, Play Jenga, Push Button, Put Rubbish in Bin,} and \textit{Slide Block to Target}. Images of the representative RLBench tasks evaluated are shown in \Cref{fig:Rep_images}. These tasks require reasoning over spatial layouts, geometric constraints, and trajectory shaping, stressing the coordination of end-effector (EE) position, orientation, and gripper actuation. Performance is reported as the task success rate, averaged over held-out seeds.

\begin{table}[t]
  \centering
  \caption{Comparison of CrossInstruct with a VLM reasoning baseline \cite{li2025hamster} and pure RL approaches (SAC, TD3) on RLBench tasks. Numbers denote success rates (higher is better).}
  \label{tab:crossinstruct_rlbench}
  \begin{adjustbox}{max width=\linewidth}
  \begin{tabular}{lrrrr}
    \toprule[1.5pt]
    \textbf{Method} & \textbf{basketball} & \textbf{peg} & \textbf{close drawer} & \textbf{slide block} \\
    \midrule
    CrossInstruct   & 0.90 & 0.25 & 0.90 & 0.90 \\
    VLM-Reasoning   & 0.00 & 0.20 & 0.45 & 0.20 \\
    Pure RL -- SAC  & 0.00 & 0.00 & 0.95 & 0.10 \\
    Pure RL -- TD3  & 0.00 & 0.00 & 0.40 & 0.00 \\
    \midrule
    & \textbf{jenga} & \textbf{lift block} & \textbf{rubbish} & \textbf{push button} \\
    \midrule
    CrossInstruct   & 0.55 & 0.95 & 1.00 & 0.95 \\
    VLM-Reasoning   & 0.00 & 0.00 & 0.00 & 0.30 \\
    Pure RL -- SAC  & 0.00 & 0.00 & 0.00 & 0.05 \\
    Pure RL -- TD3  & 0.00 & 0.00 & 0.00 & 0.00 \\
    \bottomrule[1.5pt]
  \end{tabular}
\end{adjustbox}
\end{table}
  
  %\begin{adjustbox}{max width=\linewidth}
  %  \begin{tabular}{lrrrr}
  %    \toprule[1.5pt]
  %    \textbf{Method} & \textbf{basketball} & \textbf{drawer} & \textbf{peg} & \textbf{lift block} \\
  %    \midrule
  %    CrossInstruct   & 0.90 & 0.90 & 0.25 & 0.95 \\
  %    VLM-Reasoning   & 0.00 & 0.45 & 0.20 & 0.00 \\
  %    Pure RL -- SAC  & 0.00 & 0.95 & 0.00 & 0.00 \\
  %    Pure RL -- TD3  & 0.00 & 0.40 & 0.00 & 0.00 \\
  %    \midrule
  %    & \textbf{jenga} & \textbf{button} & \textbf{rubbish} & \textbf{slide block} \\
  %    \midrule
  %    CrossInstruct   & 0.55 & 0.95 & 1.00 & 0.90 \\
  %    VLM-Reasoning   & 0.00 & 0.30 & 0.00 & 0.20 \\
  %    Pure RL -- SAC  & 0.00 & 0.05 & 0.00 & 0.10 \\
  %    Pure RL -- TD3  & 0.00 & 0.00 & 0.00 & 0.00 \\
  %    \bottomrule[1.5pt]
  %  \end{tabular}
  %\end{adjustbox}
%\end{table}

\begin{figure}[t]
\includegraphics[width=0.33\linewidth]{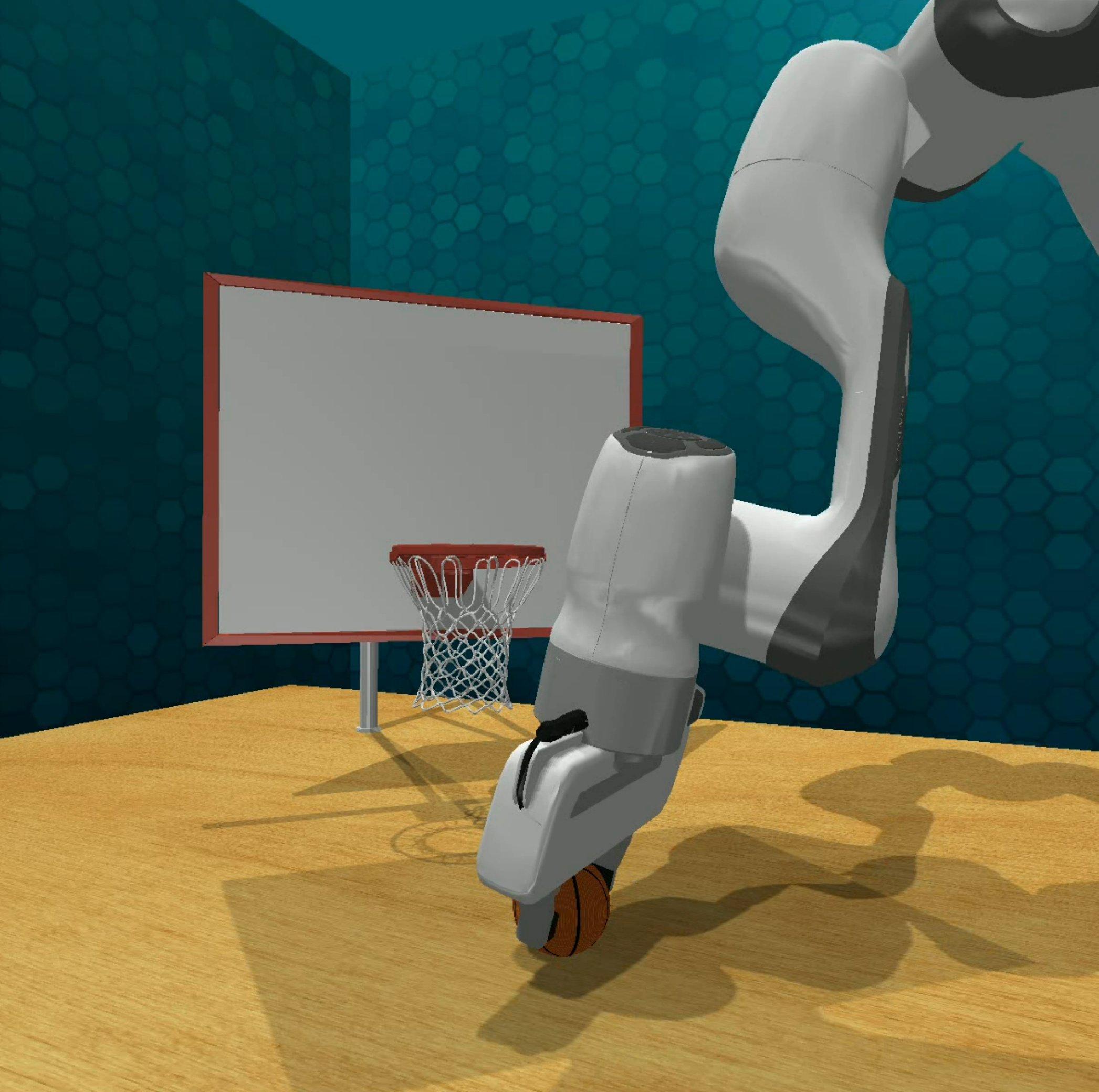}%
\includegraphics[width=0.33\linewidth]{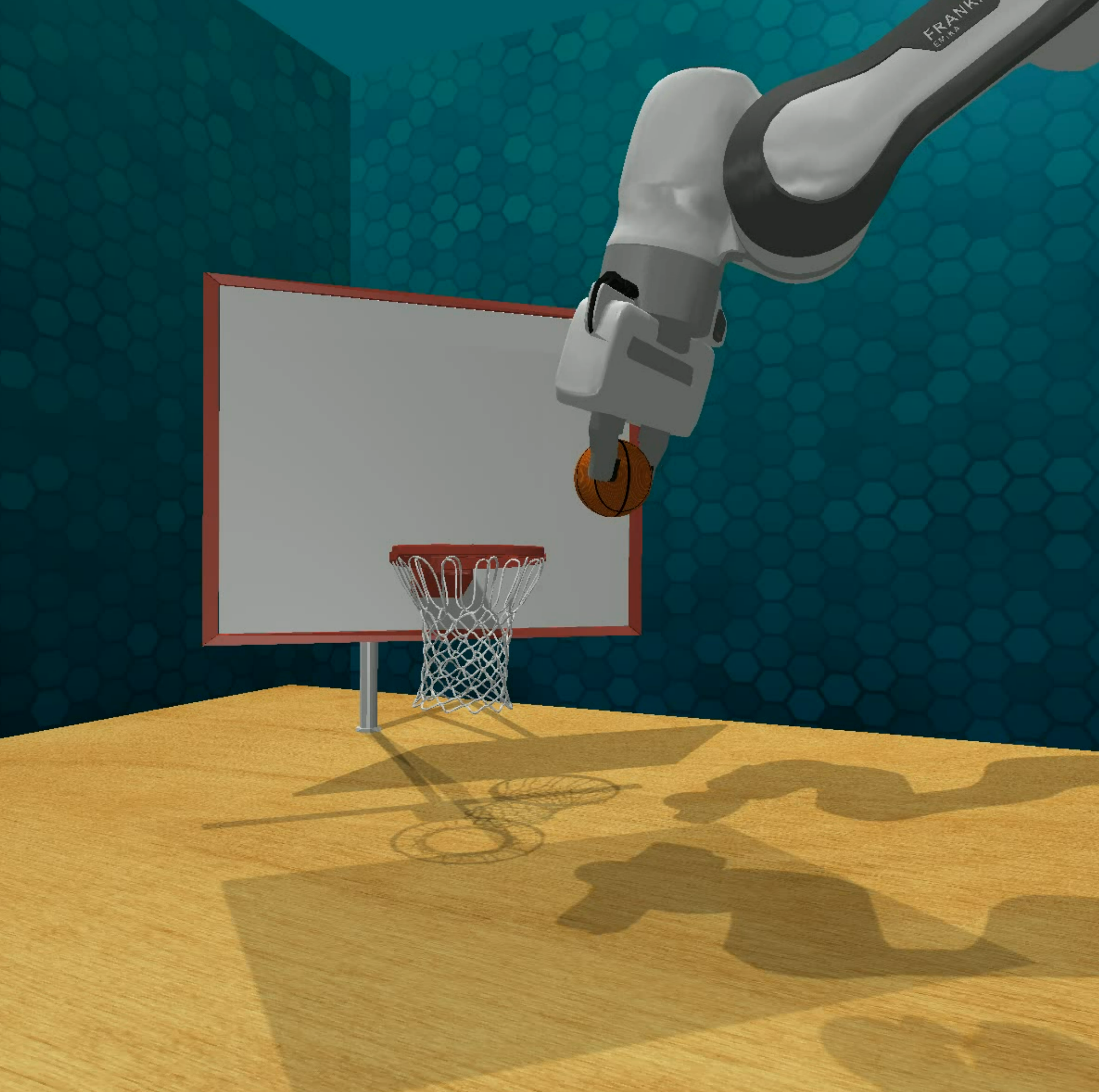}%
\includegraphics[width=0.33\linewidth]{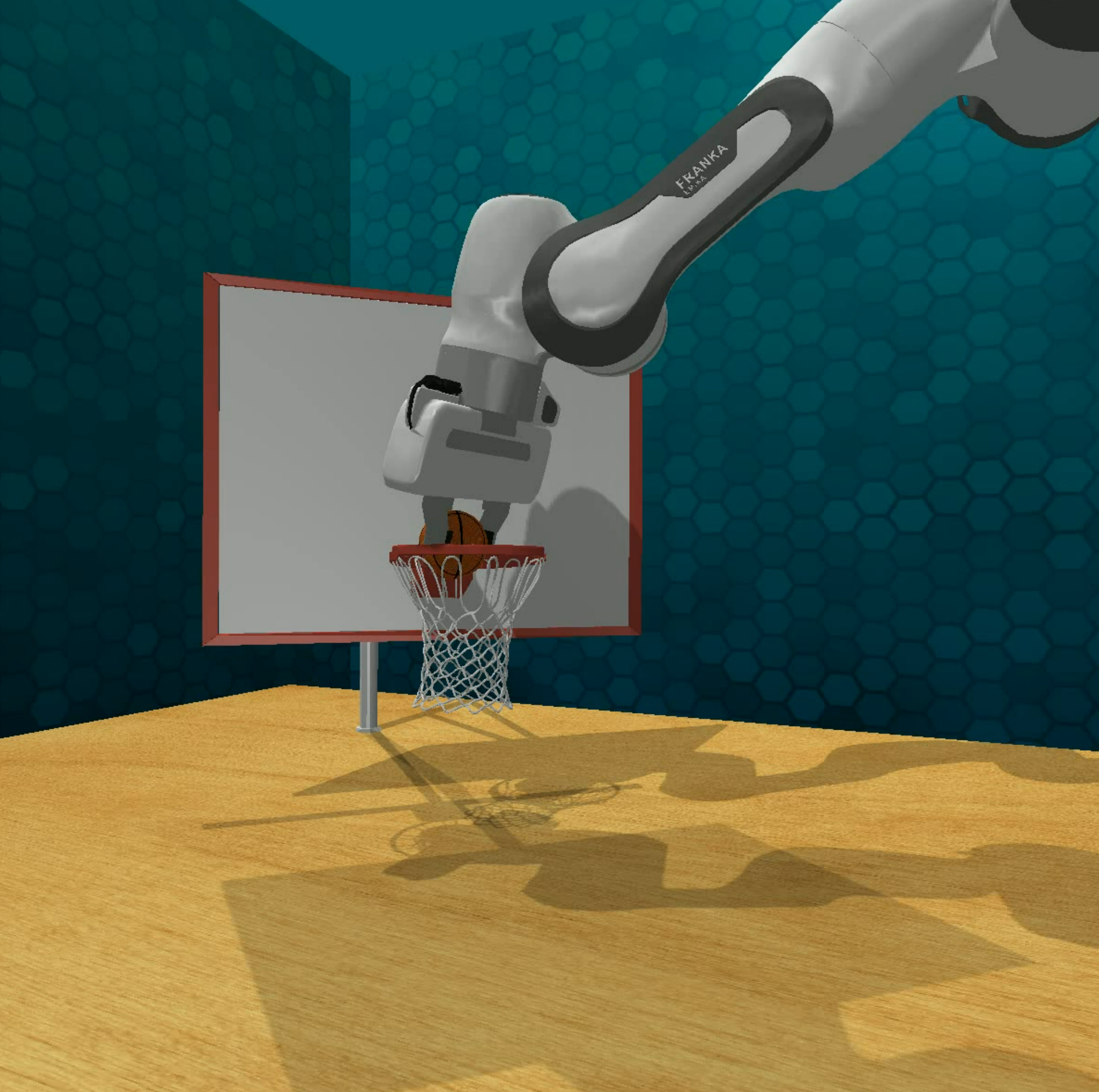}%
\caption{CrossInstruct, through hierarchical precision coupling, can generate spatially-accurate motions, enabling the robot to precisely pick up and place the basketball into the hoop.}
\label{fig:basketball_sequence}
\end{figure}

\subsection{CrossInstruct Setup Details}
For each RLBench task, we evaluate the performance of CrossInstruct to execute on cross-modal instructions, without additional fine-tuning. We collect a single sketched demonstration on one random seed. A human annotator overlays freehand geometry (curves, arrows, boundaries) and scribbled text directly on task images. The CrossInstruct reasoning model consumes this mixed-modality input, infers the intent of the task, and identifies precision-critical waypoints. For each evaluation seed, we follow the implementation details in \cref{sec: CrossInstruct} to produce an ordered 3D waypoint sequence, along with end-effector orientations and gripper actions generated by the reasoning model. The trajectory is tracked to the closest joint-space trajectory by leveraging a lower-level controller, or alternatively with reactive motion primitives \cite{RMPs, Fast_diff_int}, preserving sketch-implied trajectory shaping and collision clearance.

\subsection{Baselines and Comparisons}
To analyze the performance of CrossInstruct, we compare it against the following sets of baselines on a range of tasks from RLBench.

\textbf{VLM-Reasoning (no precision coupling):} This baseline follows the same pipeline and uses the same sketched demonstration and evaluation seeds as CrossInstruct. However, the reasoning model directly draws trajectories over the task images and outputs EE poses and gripper actions without invoking the pointing VLM. This removes pixel-level keypoint supervision, allowing us to isolate the contribution of the hierarchical precision-coupling module.

\textbf{Reinforcement Learning (RL):} We train two widely used off-policy RL algorithms: Twin Delayed Deep Deterministic Policy Gradient (TD3) \cite{Fujimoto2018AddressingFA} and Soft Actor-Critic (SAC) \cite{haarnoja2018soft}. Both agents are trained with sparse binary rewards and a budget of one million environment steps per task. TD3 is an actor–critic method that stabilizes deterministic policy gradients with clipped double Q-learning and delayed policy updates. SAC is an entropy-regularized method that learns stochastic policies to maximize exploration and robustness. We use their standard implementations without behavior cloning (BC) priors.

All baselines are evaluated on the held-out seeds. This suite of experiments allows us to (i) quantify the gains from precision coupling (CrossInstruct vs. VLM-Reasoning), and (ii) contrast instruction-driven learning with exploration-driven RL, i.e., CrossInstruct and VLM compared to TD3 and SAC.

\subsection{Real-World Setup}
To assess robustness under domain and embodiment shift, we qualitatively evaluate CrossInstruct on two real-world tabletop tasks:  

\begin{itemize}
    \item \textbf{Place Cups:} Four colored cups and four colored plates are scattered on a tabletop. The robot must match each cup with the plate of the corresponding color, while avoiding collisions and mismatches. Success requires inferring the latent color matching rule from the demonstration, selecting grasp-friendly waypoints, and executing trajectories robust to object displacement.  
    \item \textbf{Saw Block:} The robot must grasp a hand saw and perform three full back-and-forth strokes along a marked line. This task stresses tool use, orientation control along a constrained path, and correct interpretation of repetition cycles (called '3x' in the demonstration).  
\end{itemize}

For both tasks, demonstrations are collected on images captured in setups with different backgrounds, lighting, and end-effector appearances from the deployment environment. This directly tests CrossInstruct’s ability to bind abstract, sketched intent to new scenes without additional training or domain adaptation. Images of these tasks can be seen in \cref{fig:real_exp_setup}.

\begin{figure}[t]
\includegraphics[width=0.33\linewidth]{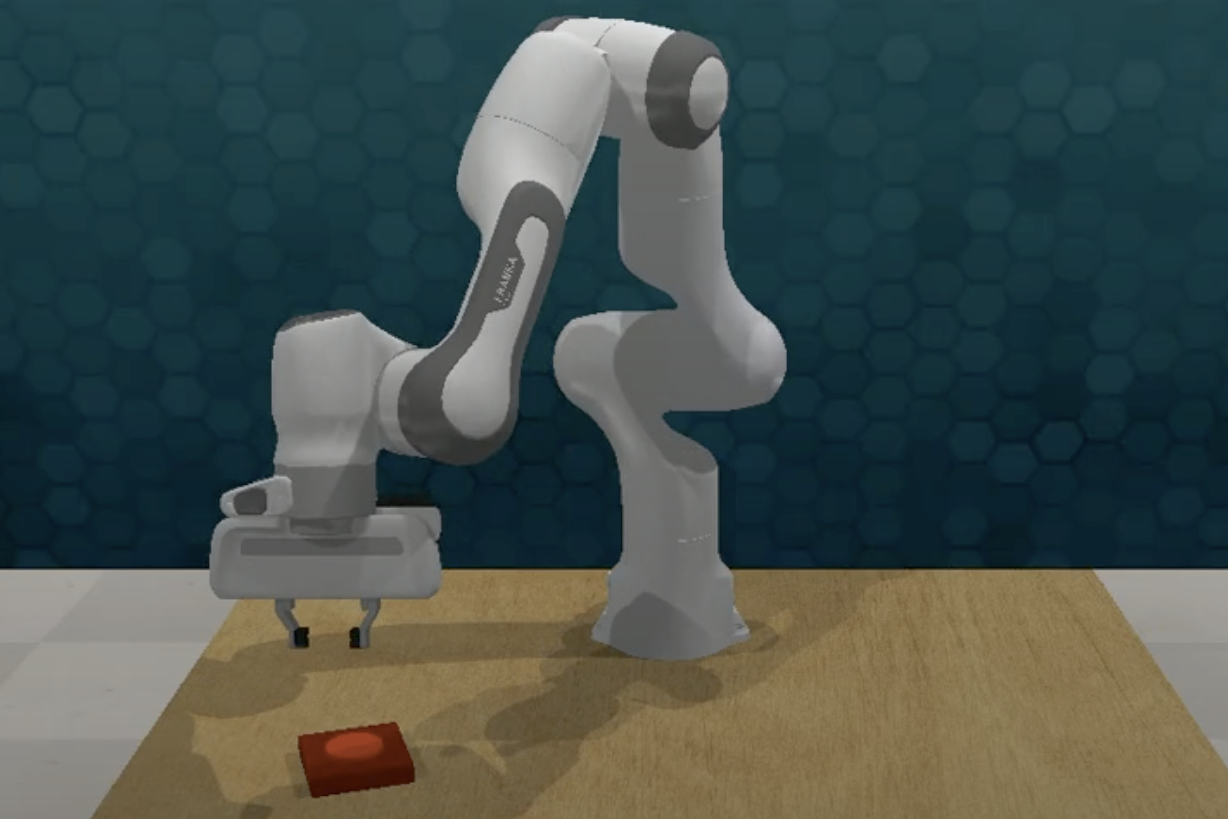}%
\includegraphics[width=0.33\linewidth]{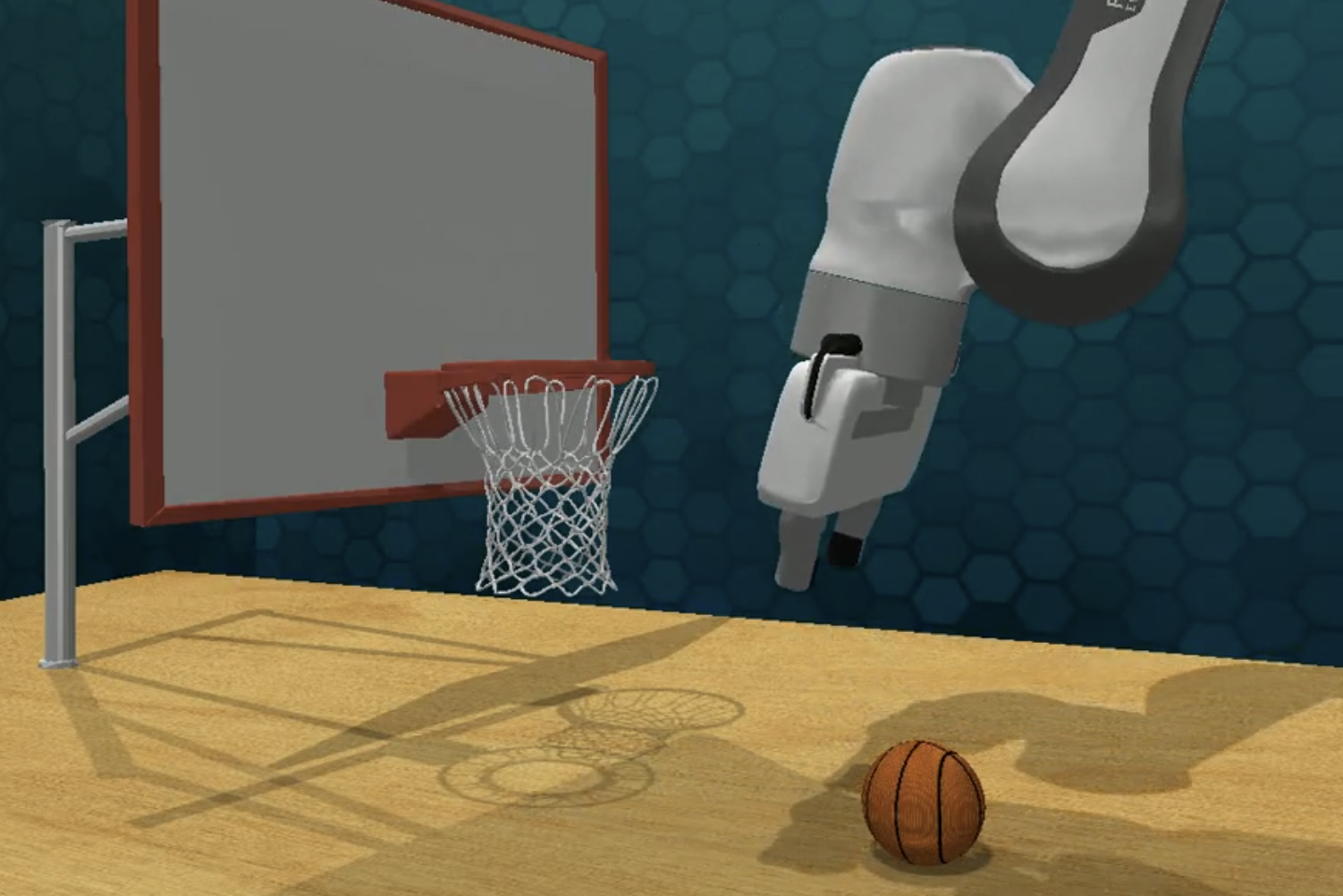}%
\includegraphics[width=0.33\linewidth]{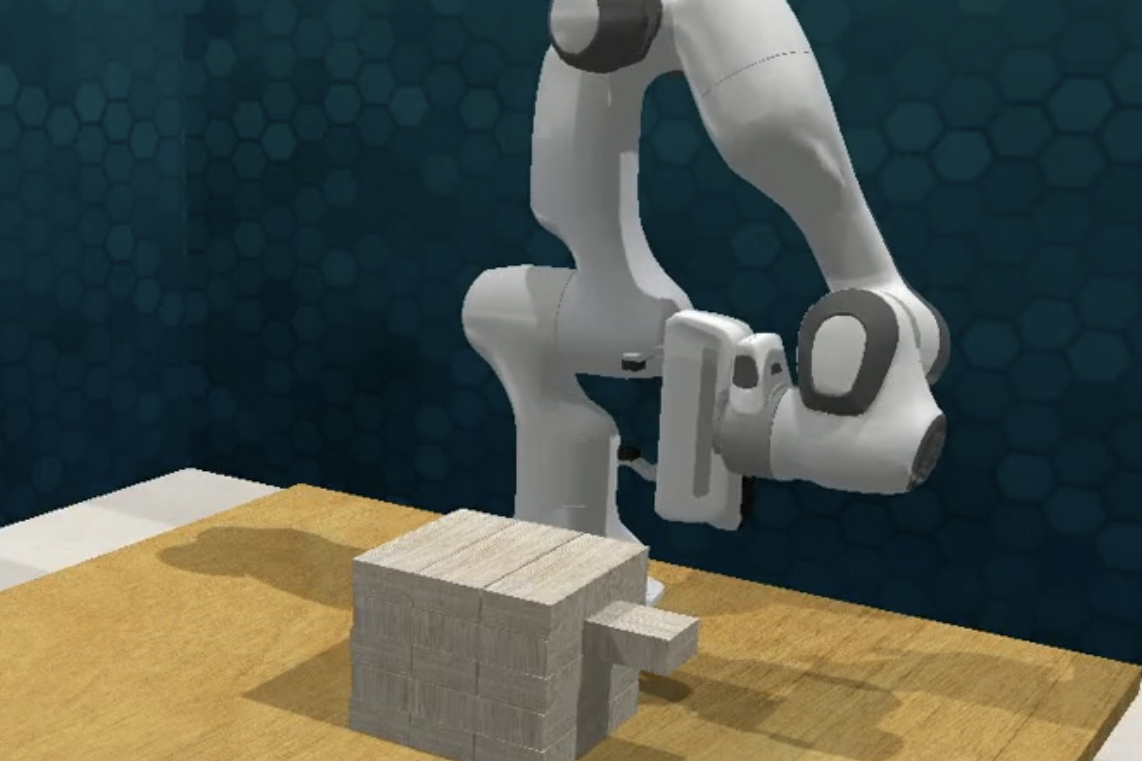}%
\caption{Without hierarchical precision coupling, directly seeking the VLM reasoning model to provide trajectories often leads to the robot not adequately reaching the object that we need to interact with, such as the button, basketball and Jenga block.}\label{fig:fail}
\end{figure}

\subsection{Results and Analysis }
\paragraph{Simulation} CrossInstruct consistently outperforms both TD3 and SAC across nearly all RLBench tasks. The gains are particularly pronounced in precision-sensitive tasks such as \textit{Basketball-in-Hoop} and \textit{Push Button}, where pixel-level keypoint grounding prevents small misalignments that would otherwise cause failure. A qualitative rollout for \emph{basketball-in-hoop} is shown in \Cref{fig:basketball_sequence}, which illustrates spatially accurate pick-and-place into the hoop enabled by hierarchical precision coupling. A notable exception is \textit{Close Drawer}, a short-horizon task with modest precision demands, where exploration-based RL methods occasionally succeed.

\begin{wrapfigure}{L}{0.48\linewidth}
\centering
    \includegraphics[width=0.49\linewidth]{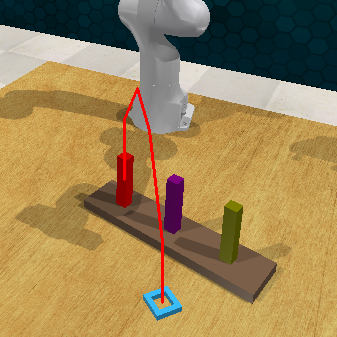}%
\includegraphics[width=0.49\linewidth]{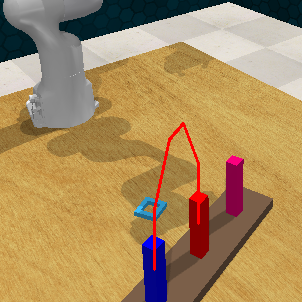}
\caption{Trajectories inferred with VLM-reasoning, without precision coupling can often mistakenly spatially-ground the objects in similar colored objects. On the right, both the peg and the square are blue, causing an erroneous trajectory (red). This does not happen in the left figure where blue pegs are not present.}\label{fig:error_exp}
\end{wrapfigure}
\begin{figure*}[t]
\centering
\framebox{%
  \parbox{0.585\linewidth}{%
    \centering
    \includegraphics[width=0.33\linewidth]{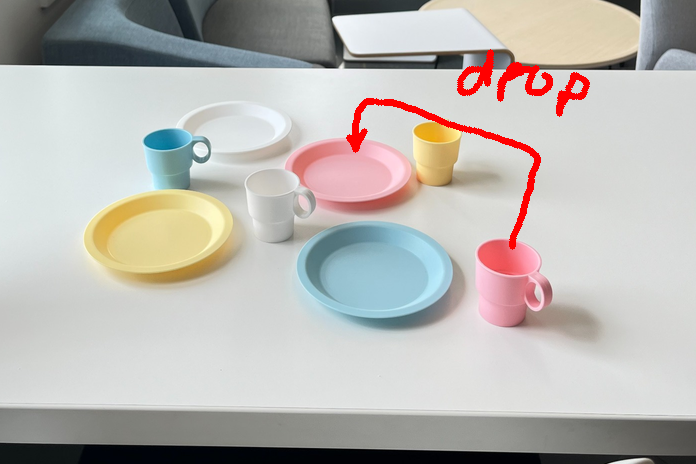}%
    \includegraphics[width=0.33\linewidth]{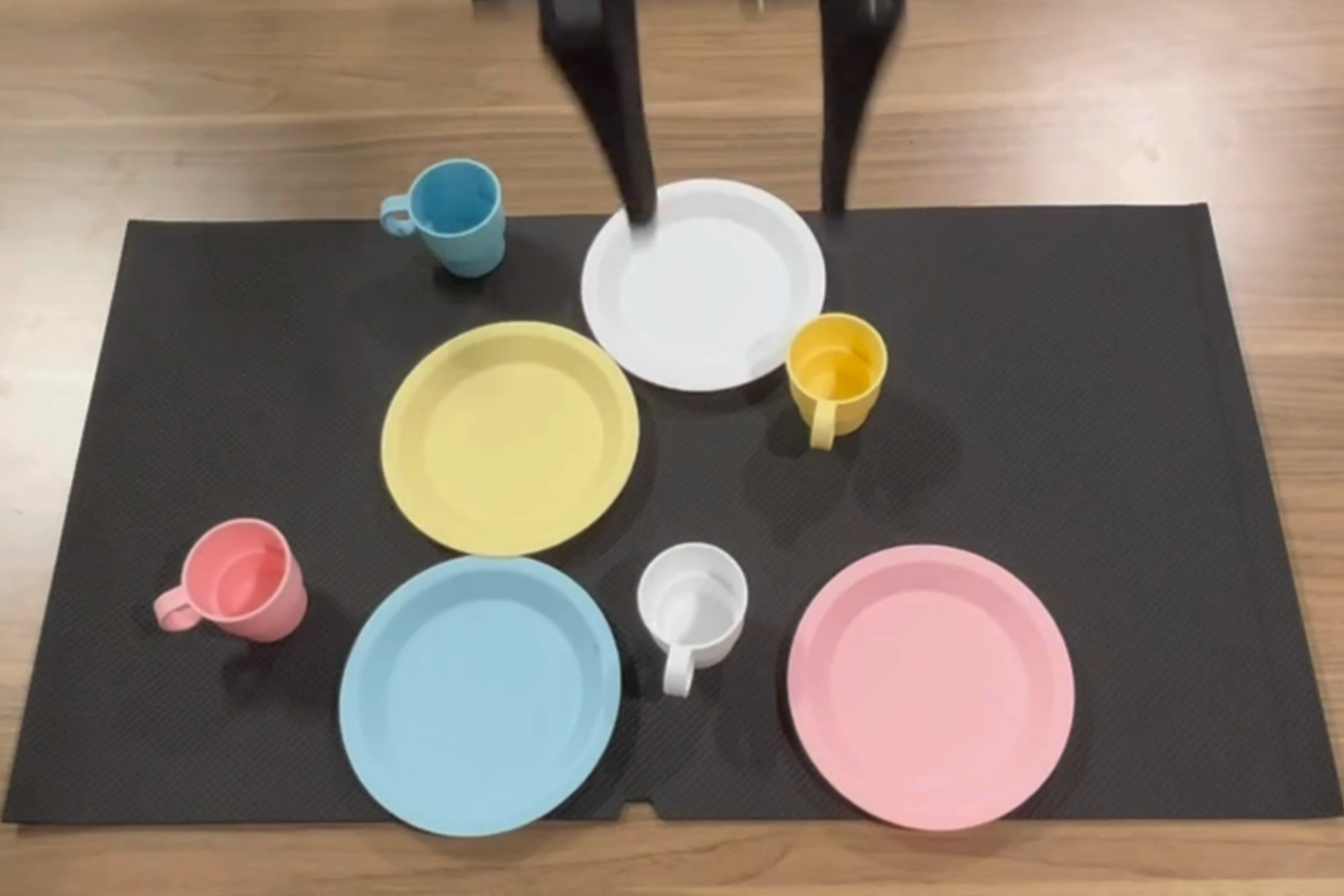}%
    \includegraphics[width=0.33\linewidth]{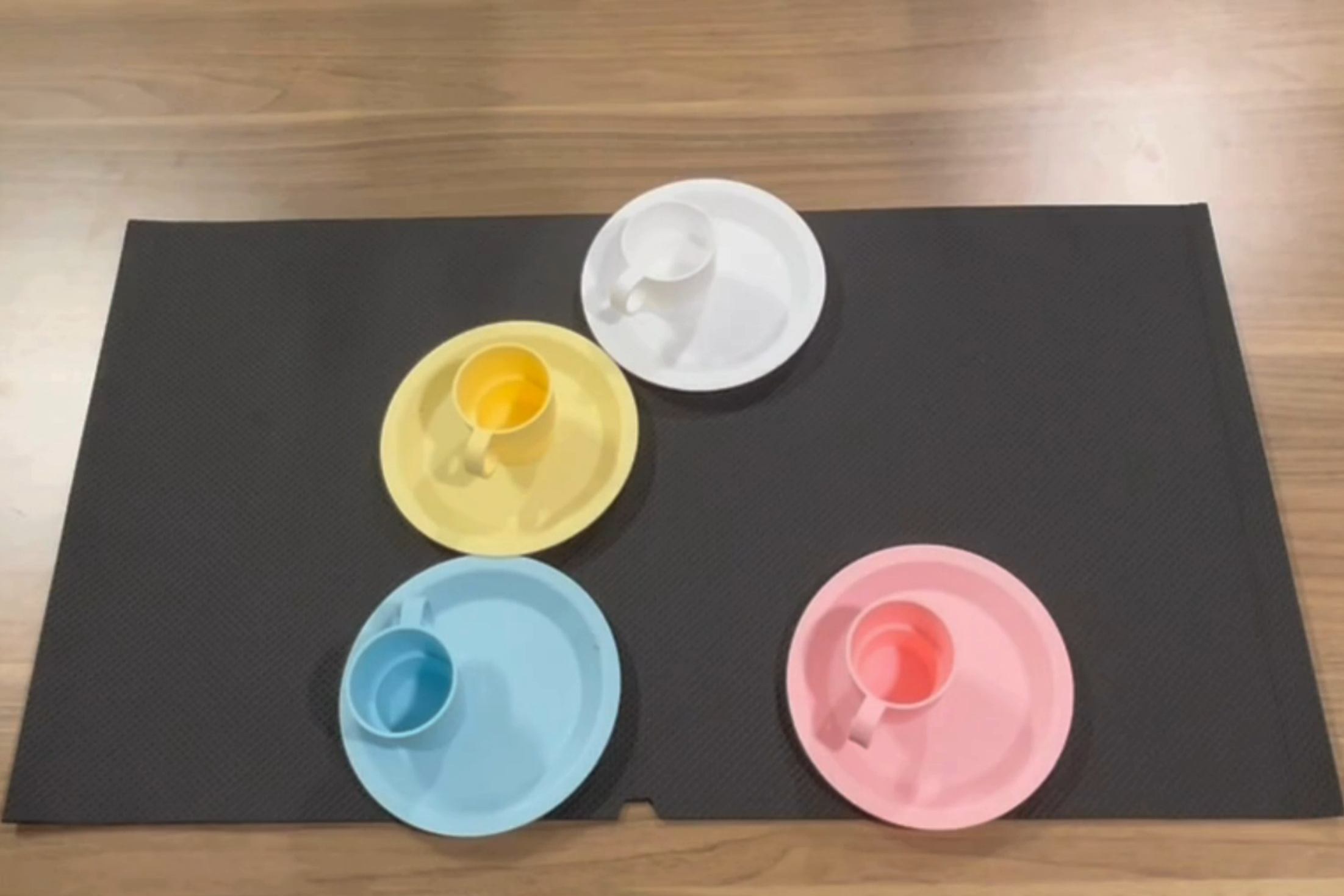}%
  }%
}
\framebox{%
  \parbox{0.39\linewidth}{%
    \centering
    \includegraphics[width=0.495\linewidth]{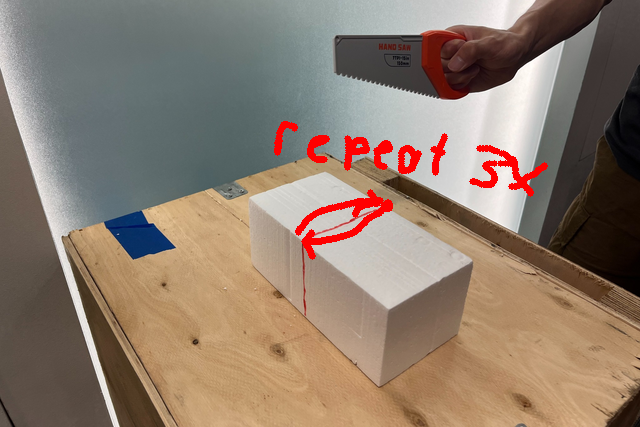}%
    \includegraphics[width=0.495\linewidth]{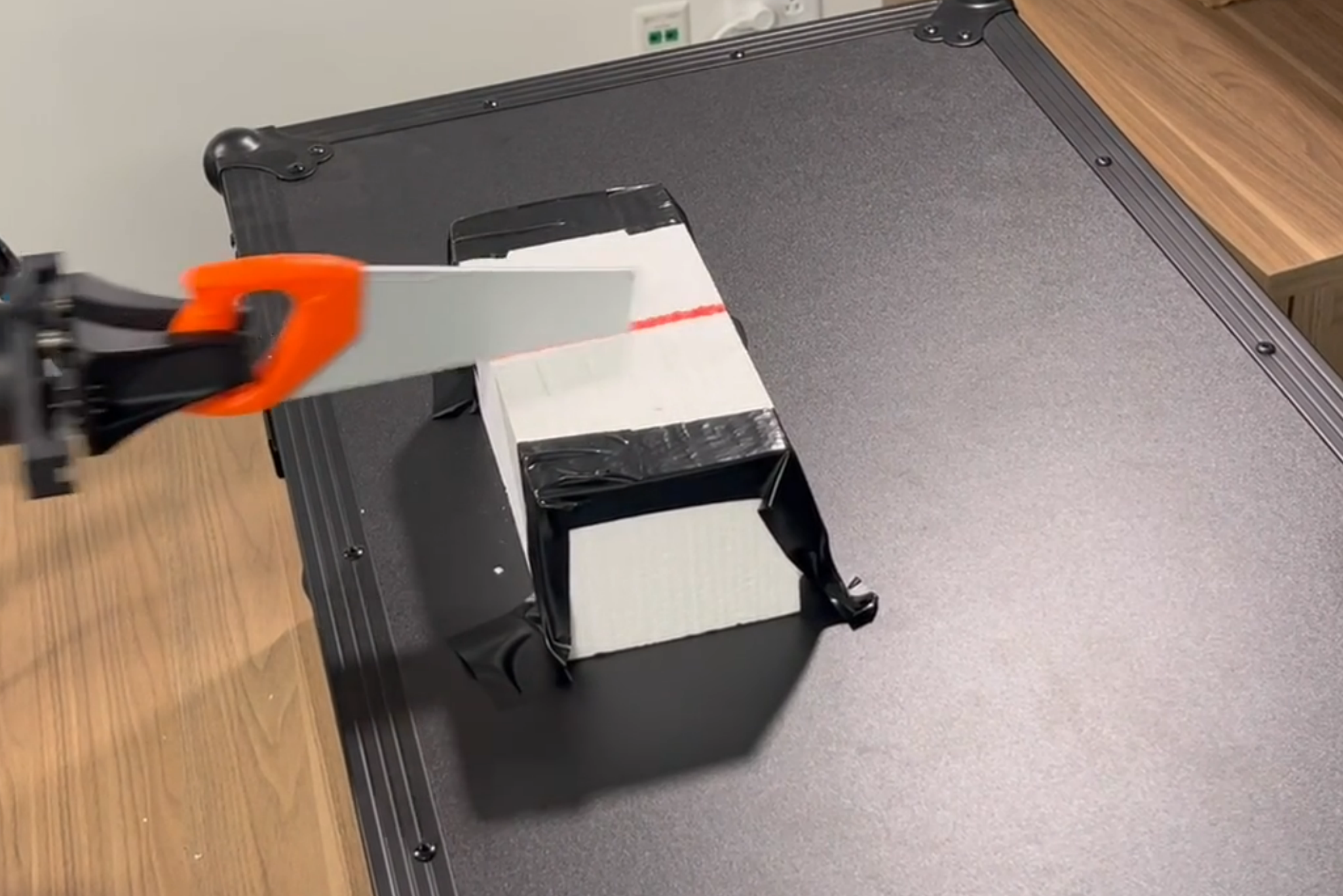}%
  }%
}
\caption{Generalizing Cross-modal instructions to new setups, with the \emph{Place Cups} task shown in the left and the \emph{Saw Block} task on the right. An image with an instruction is overlaid, and we observe that the setups that the instructions are defined over are visually different from the execution setup, highlighting the generalization exhibited. We can also provide instructions such as ``repeat 3x'' which specifies the sawing motion to be repeated three times.}\label{fig:real_exp_setup}
\vspace{-1em}
\end{figure*}

Relative to pure VLM-Reasoning, CrossInstruct demonstrates clear advantages in cluttered or low-contrast scenes, such as \textit{Basketball-in-Hoop} and \textit{Play Jenga}, where semantic sketching alone is insufficient to localize exact geometric keypoints. Interestingly, for \textit{square block on peg}, the reasoning-only variant performs comparably, as the bright, simple visual cues make peg–hole delineation achievable. However, when there exists multiple objects with similar colors, such as the setup illustrated in \cref{fig:error_exp}, the absence of the smaller pointing model can lead to erroneous results. The VLM-reasoning baseline occassionally fails due to the spatial coordinates being inaccurate, and the robot not reaching desired poses, with notable examples in \cref{fig:fail}. These findings suggest that precision coupling is most valuable when object boundaries are ambiguous, small, or occluded, and when success depends on the need to generate relatively precise motions.

\paragraph{Real World} CrossInstruct generalizes robustly to physical setups, successfully executing mixed sketch–text instructions despite significant differences in visual appearance and manipulator kinematics. In \textit{Place Cups}, the system correctly infers the color-matching rule, plans grasp-friendly motions, and executes without collisions. In \textit{Saw Block}, it respects the “repeat 3x” instruction, maintaining orientation along the constrained sawing path. Both tasks are completed without depth sensing or closed-loop replanning, highlighting CrossInstruct’s ability to bind abstract intent to real-world dynamics. Example cross-modal instructions, along with the generalized setup, which differs from the environment in which the instructions were given, are shown in \cref{fig:real_exp_setup}.

Overall, the results demonstrate that CrossInstruct effectively combines high-level semantic reasoning with low-level geometric precision, yielding robust instruction-following policies that generalize across tasks, seeds, and embodiments.

\begin{figure}[t]
\centering
\includegraphics[width=0.495\linewidth]{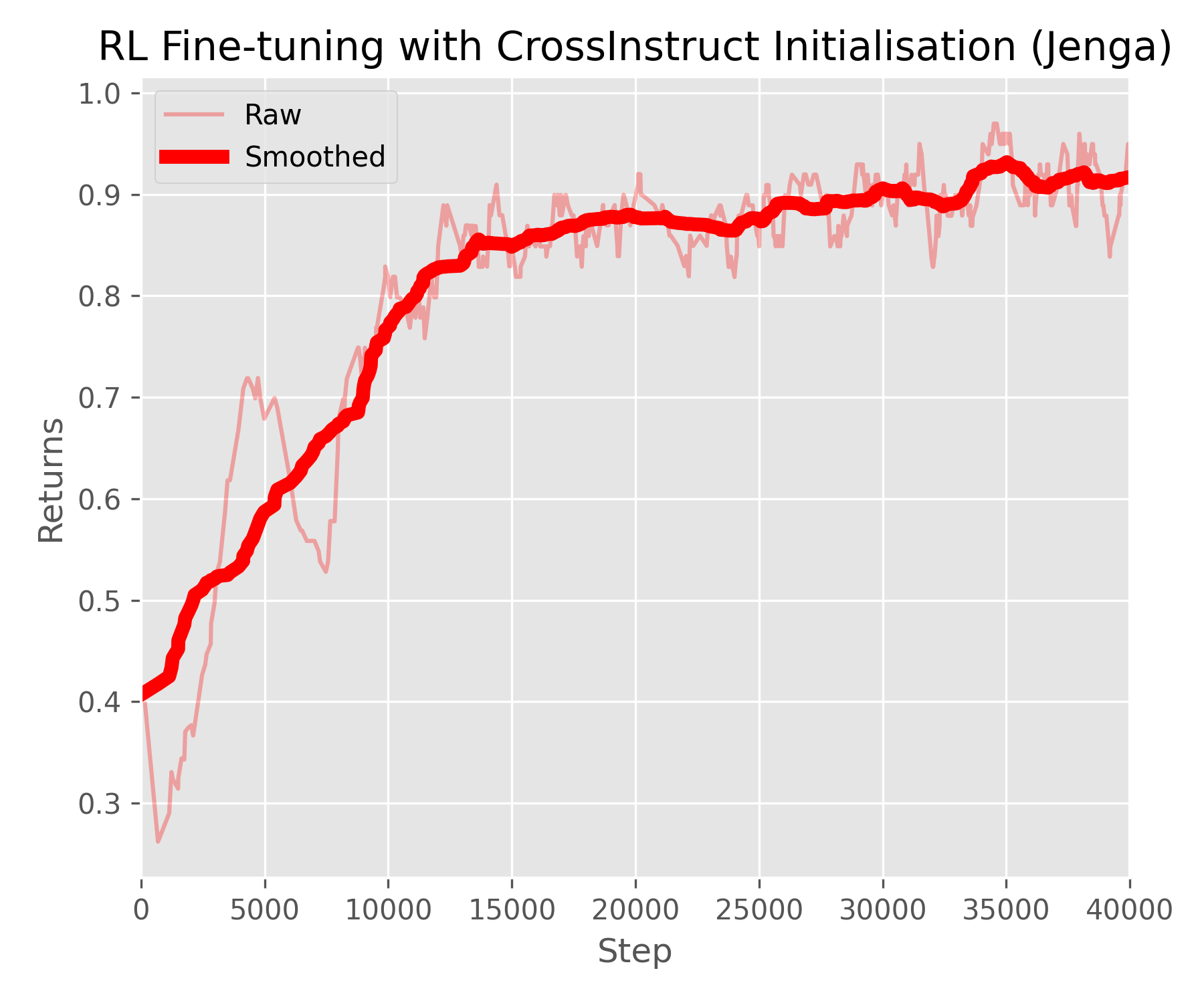}%
\includegraphics[width=0.495\linewidth]{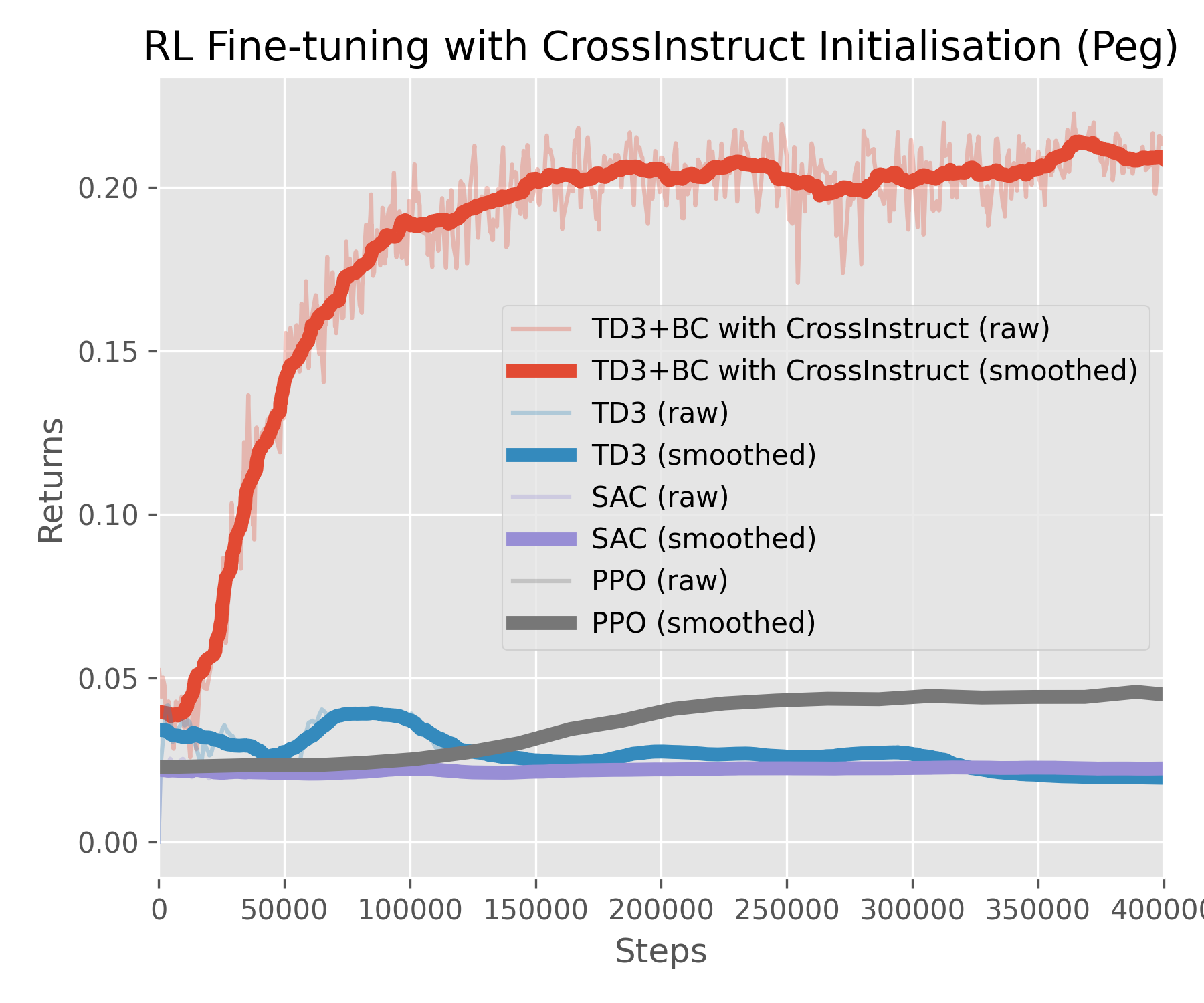}
\caption{We sample trajectories from CrossInstruct to initialize policies for RL training. Returns against the number of steps shown (left: Jenga; right: Peg). As the Jenga task uses a sparse binary reward, training from scratch gives no nonzero returns. For the Peg task, we illustrate the performance of baseline RL methods trained from scratch against our TD3+BC approach. We observe that CrossInstruct provides an effective generative model to initialize RL training.}\label{fig: RL_comp}
\end{figure}

\subsection{CrossInstruct as a Generative Model For Reinforcement Learning Initialization}

While CrossInstruct produces executable trajectories, the broader advantage lies in its ability to serve as a \emph{generative model} to initialize and improve downstream reinforcement learning (RL). By sampling from the trajectory distribution $p(\tau \mid \mathcal{I}, \mathcal{V}, \mathcal{P})$, we obtain diverse, yet semantically consistent, rollouts that can be directly integrated into a reinforcement learning (RL) pipeline. We then follow the TD3+BC approach outlined in \cref{subsec:refine}, allowing the trajectories to provide additional regularization.

We evaluated this generative initialization on two representative RLBench tasks: \emph{Play Jenga} and \emph{Insert Square Block on Peg}. Both of these tasks are difficult and require precise motions. We run $40$k RL steps for the \emph{Jenga} task, and $400k$ for the \emph{Peg} tasks. We additionally train policies on these tasks from scratch using comparison approaches TD3 \cite{Fujimoto2018AddressingFA}, SAC \cite{haarnoja2018soft}, and PPO \cite{schulman2017ppo}. In the Jenga task, the reward is designed as a binary ``success'' or ``failure'' signal, and our CrossInstruct-initialized policies reliably learn to dislodge the correct block, converging to a success rate around $90\%$, while training from scratch cannot produce a nonzero return. In the peg-insertion task, CrossInstruct significantly accelerates learning, whereas the baseline methods cannot reach comparable returns, even after 400k steps. The returns for both the Jenga and Peg tasks are illustrated in \cref{fig: RL_comp}. This shows that CrossInstruct not only produces executable plans, but also acts as a low-cost trajectory generator to reduce sample complexity in reinforcement learning. Crucially, stochastic sampling from $p(\tau)$ injects natural variability in the demonstrations, improving the robustness of the policy. 

\textbf{Experiments Summary:} In eight RLBench tasks, CrossInstruct substantially outperforms both a VLM-reasoning baseline (without precision coupling) and pure RL (SAC/TD3) trained from scratch, confirming the benefit of hierarchical precision coupling and multi-view lifting. In both simulated and real-world settings, CrossInstruct executes mixed sketch–text instructions without fine-tuning, despite substantial differences in appearance, correctly binding abstract intent, such as color matching and ``repeat 3x'' tool use, to robust motions. Additionally, sampling trajectories from $p(\tau)$ provides scalable demonstrations that both initialize and regularize RL, enabling training of robust downstream policies.

\section{Conclusions and Future Work}
In this work, we introduce \emph{CrossInstruct}, a new paradigm for teaching robots through \emph{cross-modal instructions}. These are rough sketches and textual labels provided directly on images of an environment, which can be different from the workspace of the robot. This representation avoids the high cost of collecting kinesthetic or teleoperated demonstrations while still conveying rich semantic and geometric information about the intended behavior. Our framework achieves this by introducing a hierarchical precision coupling module that integrates a large reasoning vision–language model with a lightweight, fine-tuned pointing model. Extensive evaluations in both simulation and real hardware highlight the effectiveness and generalization capacity of the approach. Beyond execution, we showed that CrossInstruct acts as a generative model for robot trajectories, enabling effective downstream reinforcement learning. 

A key avenue for future inquiry is to enable humans to provide cross-modal information to correct the robot's behavior after executing motions. Here, the human will be in the loop viewing videos of robot execution and providing cross-modal instructions to iteratively shape and correct the robot's behavior. Additionally, integrating large pre-trained models, such as DUSt3R \cite{DUSt3R_cvpr24}, VGER \cite{VGER}, or VGGT \cite{wang2025vggt}, to enable synthetic camera views from novel camera poses, thereby reducing the burden to physically position the cameras.
\bibliographystyle{IEEEtran}
\bibliography{ref}

\end{document}